\PassOptionsToPackage{table,xcdraw}{xcolor}
\PassOptionsToPackage{normalem}{ulem}
\documentclass{jingdong}

\usepackage{amsmath,amsfonts,bm}

\def\eqref#1{equation~\ref{#1}}

\def\1{\bm{1}}

\DeclareMathAlphabet{\mathsfit}{\encodingdefault}{\sfdefault}{m}{sl}
\SetMathAlphabet{\mathsfit}{bold}{\encodingdefault}{\sfdefault}{bx}{n}

\usepackage{hyperref}
\usepackage{url}

\usepackage{booktabs}               
\usepackage{makecell}               
\usepackage{graphicx}               
\usepackage{pifont}                 
\usepackage{multirow}
\usepackage{caption}
\usepackage{wrapfig}

\usepackage{tcolorbox}
\tcbuselibrary{breakable,skins}

\definecolor{headerblue}{RGB}{41,84,144}
\definecolor{rowblueA}{RGB}{237,243,250}
\definecolor{rowblueB}{RGB}{255,255,255}
\definecolor{highlightblue}{RGB}{205,224,246}

\newcommand{\cmark}{\textcolor{teal}{\ding{51}}}
\newcommand{\xmark}{\textcolor{gray}{\ding{55}}}

\fancypagestyle{firststyle}{
  \fancyhf{}
  \fancyhead[L]{%
    \vspace*{2mm}%
    \includegraphics[width=130pt]{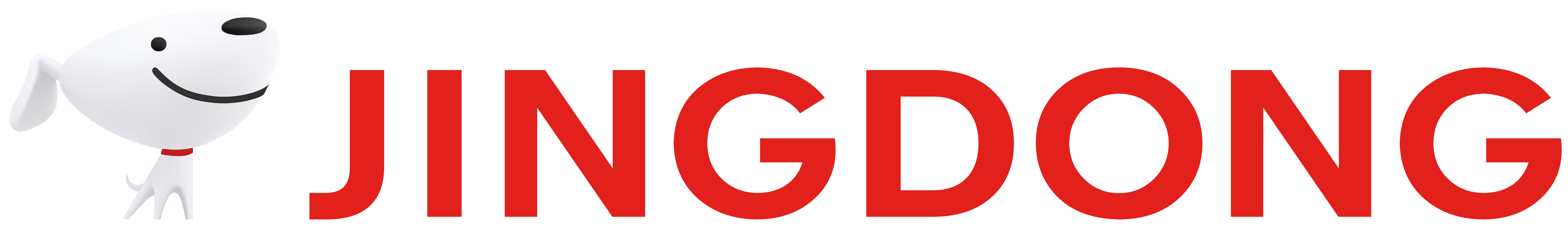}%
  }
  \fancyfoot[C]{\thepage}

}
\title{When Non-Adversarial Noise Hurts Learning: \\ Benchmarking Skill Self-Evolution in Agents}
\author[1,2,3,*,\S]{Tao~Yu}
\author[2,*]{Zhongtian~Luo}
\author[2,*]{Yuchen~Long}
\author[2]{Yiheng~Lu}
\author[5]{Yuxuan~Zhou}
\author[2]{Yanpei~Gong}
\author[2]{Siyi~Cao}
\author[2]{Yuqing~Wu}
\author[2]{Haopeng~Jin}
\author[3]{Zhenghao~Zhang}
\author[3]{Yujia~Yang}
\author[1,\S]{Jingmin~Zhu}
\author[1,\S]{Yunyun~Hou}
\author[2,3]{Xinming~Wang}
\author[3]{Hongzhu~Yi}
\author[1,\ddagger]{Jiachen~Liu}
\author[1]{Sirui~Wang}
\author[1]{Nan~Duan}
\author[2,3,\ddagger]{Yan~Huang}
\author[2,3]{Liang~Wang}
\author[4,\ddagger]{Tianjin~Huang}
\author[1,\ddagger]{Yanyan~Zou}

\affiliation[1]{JD Future Academy}
\affiliation[2]{CASIA}
\affiliation[3]{UCAS}
\affiliation[4]{University of Exeter}
\affiliation[5]{Tsinghua University}

\date{Oct. 14, 2026}
\checkdata[Code]{\url{https://github.com/yutao1024/RSEClaw}}
\checkdata[Data]{\url{https://huggingface.co/datasets/Kirito-Lab/RSE-Bench}}

\abstract{
Large language model agents increasingly improve by distilling experience into reusable skills, but errors in that experience can persist in learned rules and affect later tasks. Existing evaluations largely examine clean learning or perturbations encountered after skills are frozen, leaving robustness during self-evolution insufficiently understood. We introduce RSE-Bench, to our knowledge the first benchmark for evaluating skill self-evolution under ordinary, non-adversarial noise. It spans four domains and six benchmarks, with controlled perturbations to instructions, environmental evidence, execution records, and outcome--record pairings that preserve the underlying tasks and native scoring rules. The protocol separates noise during learning from input-side noise at evaluation, testing frozen skill stores on held-out tasks. We also introduce \textsc{RSEClaw}, a method-agnostic framework supporting consistent noise injection and process diagnostics of actual exposure and skill updates. Across three methods and two backbones, noisy learning retains only $25.5\%$--$42.1\%$ of the aggregate gain achieved by clean self-evolution, depending on the noise type. Perturbations confined to execution records and outcome--record pairings cause average losses comparable to input-side noise, despite leaving the original executions and evaluations intact. Diagnostics further identify differences in candidate skill text that coexist with small changes in mean held-out performance.

}

\begin{document}
\addtocontents{toc}{\protect\setcounter{tocdepth}{-1}}
\maketitle

\begin{NoHyper}
\renewcommand{\thefootnote}{*}
\footnotetext{Equal contribution.}
\renewcommand{\thefootnote}{\S}
\footnotetext[0]{Work done during an internship at JD Future Academy.}
\renewcommand{\thefootnote}{\textdaggerdbl}
\footnotetext{Corresponding author.}
\end{NoHyper}

\section{Introduction}

Large language model (LLM) agents increasingly improve themselves without weight updates: after completing tasks, they distill execution experience into persistent, reusable artifacts such as skill libraries \citep{voyager2023, skillweaver2025}, natural-language insights \citep{expel2024, 202609.0154}, workflow memory \citep{awm2024}, and evolving playbooks \citep{ace2025, suzgun2026dynamic}. Such skill self-evolution can turn solved tasks into assets for future ones, and recent methods report substantial gains over skill-free baselines across coding, web, and office domains \citep{skillopt2026, rethinkskill2026}.

These gains are typically measured without systematically perturbing the experience used for learning. Deployed agents face redundant instructions, irrelevant and near-duplicate files, fragmented logs, and outcomes paired with other attempts' records. For static agents, prior work primarily measures the execution-time cost of such noise \citep{agentnoisebench2026, gsmic2023}. For self-evolving agents, an error can also persist beyond the episode. An agent might misread an outdated worksheet and distill that experience into an unreliable rule for choosing templates, affecting later clean tasks that reuse it. This raises an additional question: whether noise makes the agent learn the wrong thing and carry it forward.

\begin{figure*}[t]
    \centering
    \vspace{-15pt}
    \includegraphics[width=0.98\textwidth]{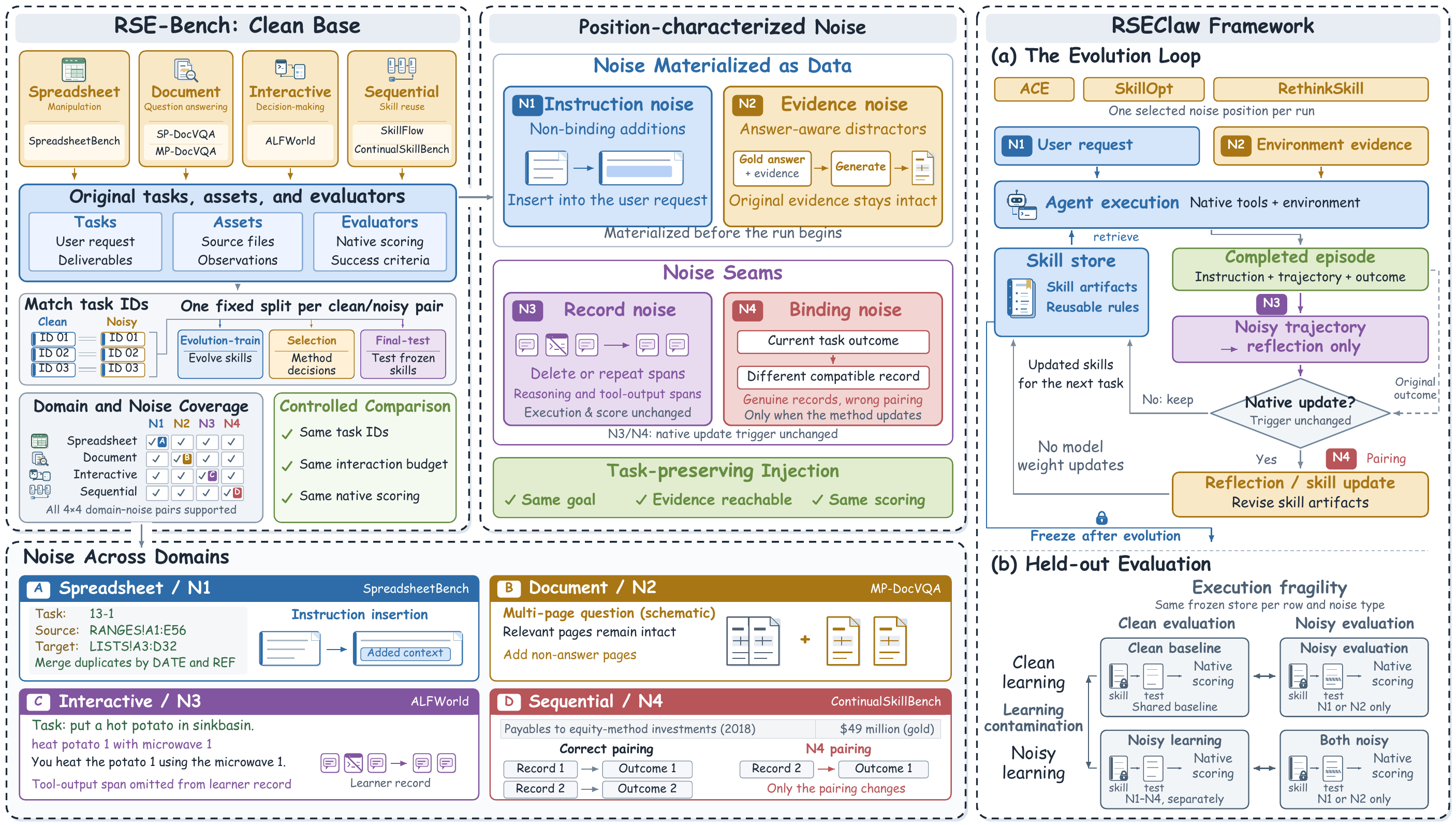}
    \caption{Overview of RSE-Bench and \textsc{RSEClaw}: clean tasks from six sources, noise injected at four entry points of the learning loop, and evaluation on held-out tasks with the skill store frozen.}
    \label{fig:overview}
    \vspace{-15pt}
\end{figure*}

Existing evaluations do not directly isolate this effect. Noise benchmarks for agents inject perturbations at execution time and evaluate static policies, without updating skills from the perturbed episodes \citep{agentnoisebench2026, stressweb2026}. Robustness probes for evolved skills perturb inputs after evolution completes, with the skills frozen \citep{rethinkskill2026}. A parallel security literature studies malicious corruption of agent memory and self-evolution, focusing on attack success or safety degradation \citep{skillmisevo2026, memevobench2026, agentpoison2024}. These lines of work leave open a controlled evaluation of how ordinary, non-adversarial noise entering the learning loop affects the reliability of the resulting skills.

We present RSE-Bench, a benchmark for noise-robust skill self-evolution, together with \textsc{RSEClaw}, the framework that runs it. Two design decisions organize the benchmark. First, noise is characterized by its entry point: task instructions (N-Inst), execution-time evidence (N-Evid), the record visible to the learner (N-Rec), and outcome--record pairings (N-Bind). Each injection obeys task-preserving constraints that keep task goals, correct-evidence reachability, ground-truth answers, and native scoring rules unchanged. Second, learning and evaluation noise are separated: N-Inst and N-Evid can vary in either phase, while N-Rec and N-Bind apply only during learning. Clean evaluation measures the downstream effects of noisy learning; noisy evaluation of cleanly learned skills measures execution-time sensitivity. Crossing the applicable conditions further tests whether learning under noise prepares an agent for the same noise at deployment.

RSE-Bench covers four domains and six benchmarks: spreadsheet manipulation, document understanding, interactive decision-making, and skill reuse along ordered task streams. Each uses fixed evolution-train, selection, and final-test splits, with task identities matched across conditions. The final-test split is never used for learning and is evaluated with the skill store frozen. N-Rec and N-Bind transform learning experience at run time, so task variants alone cannot implement them. \textsc{RSEClaw} decouples methods from execution, scoring, and noise injection through a fixed interface. Methods implementing this interface run without condition-specific modifications. Actual exposure depends on what each method consumes, so diagnostics track noise exposure, learning activity, and selected skill versions, supported by optimizer-request logs and paired-play audits(Appendix~\ref{app:setup_audit}).

Our main experiments evaluate ACE, SkillOpt, and RethinkSkill with GPT-5.2 and DeepSeek-V4.1-Flash. Clean evolution improves the six-benchmark macro-average by $2.6$--$6.5$ percentage points, yet the four noisy-learning conditions retain only $25.5\%$--$42.1\%$ of this gain when aggregated across method--backbone pairs. Evidence noise produces the largest mean loss in both phases. Noise in execution records and outcome--record pairings causes average losses comparable to input-side noise, despite leaving the original execution and evaluation intact. In an ACE ablation on SpreadsheetBench with DeepSeek-V4.1-Flash, learning under the same noise used at evaluation also fails to improve performance over clean learning. Our contributions are:

\textbf{A benchmark for noise-robust skill self-evolution.}
To our knowledge, RSE-Bench is the first benchmark evaluating skill self-evolution under ordinary, non-adversarial noise, with paired conditions separating the effects of noisy learning from execution-time sensitivity.

\textbf{The first controlled noise-injection protocol for skill self-evolution.} To our knowledge, this is the first task-preserving protocol jointly covering non-adversarial noise in instructions, evidence, execution records, and outcome--record pairings within the skill self-evolution loop. Instruction and evidence noise are materialized as data; record and binding noise operate at run time.

\textbf{\textsc{RSEClaw}, a method-agnostic evaluation framework.} A fixed interface supports consistent execution, scoring, and noise injection across methods, while process diagnostics and paired-replay audits assess actual noise exposure and learning activity. We release the framework, benchmark data, and noise generators.

\begin{table*}[t]
\centering
\vspace{-15pt}
\caption{Comparison of RSE-Bench with existing benchmarks. \textbf{Self-Evolution}: the benchmark evaluates agents that update persistent skills or memory from experience. \textbf{Injected}: controlled perturbations are injected. \textbf{Into Learning Loop}: injected noise can be distilled into the agent's persistent skills, rather than affecting only a single episode or frozen artifacts. \textbf{Non-Adversarial}: noise models ordinary deployment imperfections rather than attacks. \textbf{Task-Preserving}: injections keep tasks solvable, with ground truth and scoring unchanged. \textbf{Injection Positions}: number of distinct injection channels, per each paper's own definition. \textbf{Clean Held-out Eval}: evolved skills are evaluated on held-out tasks untouched by learning. \textbf{Pluggable Methods}: multiple self-evolution methods run behind a unified interface. \textbf{Metric}: capability (Cap.) or attack success rate (ASR).}
\label{tab:benchmark_comparison}
\resizebox{\textwidth}{!}{
\rowcolors{3}{rowblueA}{rowblueB}
\begin{tabular}{lccccccccc}
\toprule

\hiderowcolors
\rowcolor{headerblue}
 & & \multicolumn{4}{c}{\color{white}\small\textbf{Noise Injection}} & & & & \\
\rowcolor{headerblue}
\multirow{-2}{*}{\color{white}\small\textbf{Benchmark}} &
\multirow{-2}{*}{\color{white}\small\textbf{\makecell{Self-\\Evolution}}} &
{\color{white}\scriptsize Injected} &
{\color{white}\scriptsize\makecell{Into Learning\\Loop}} &
{\color{white}\scriptsize\makecell{Non-\\Adversarial}} &
{\color{white}\scriptsize\makecell{Task-\\Preserving}} &
\multirow{-2}{*}{\color{white}\small\textbf{\makecell{Injection\\Positions}}} &
\multirow{-2}{*}{\color{white}\small\textbf{\makecell{Clean\\Held-out Eval}}} &
\multirow{-2}{*}{\color{white}\small\textbf{\makecell{Pluggable\\Methods}}} &
\multirow{-2}{*}{\color{white}\small\textbf{Metric}} \\
\showrowcolors

\midrule

GSM-IC~\citep{gsmic2023} & \xmark & \cmark & \xmark & \cmark & \cmark & 1 & \xmark & \xmark & Cap. \\
AgentNoiseBench~\citep{agentnoisebench2026} & \xmark & \cmark & \xmark & \cmark & \cmark & 2 & \xmark & \xmark & Cap. \\
StressWeb~\citep{stressweb2026} & \xmark & \cmark & \xmark & \cmark & \xmark & 3 & \xmark & \xmark & Cap. \\
AgentPoison~\citep{agentpoison2024} & \xmark & \cmark & \xmark & \xmark & \xmark & 1 & \xmark & \xmark & ASR \\
MINJA~\citep{minja2025} & \cmark & \cmark & \cmark & \xmark & \xmark & 1 & \cmark & \xmark & ASR \\
MemEvoBench~\citep{memevobench2026} & \cmark & \cmark & \cmark & \xmark & \xmark & 3 & \xmark & \xmark & ASR \\
SkillMisevo-Bench~\citep{skillmisevo2026} & \cmark & \cmark & \cmark & \xmark & \xmark & 1 & \cmark & \xmark & ASR \\
StreamBench~\citep{streambench2024} & \cmark & \xmark & -- & -- & -- & -- & \xmark & \cmark & Cap. \\
LifelongAgentBench~\citep{lifelongagentbench2025} & \cmark & \xmark & -- & -- & -- & -- & \xmark & \xmark & Cap. \\
Evo-Memory~\citep{evomemory2026} & \cmark & \xmark & -- & -- & -- & -- & \xmark & \cmark & Cap. \\
SkillFlow~\citep{skillflow2026} & \cmark & \xmark & -- & -- & -- & -- & \xmark & \xmark & Cap. \\
SkillLearnBench~\citep{skilllearnbench2026} & \cmark & \xmark & -- & -- & -- & -- & \xmark & \cmark & Cap. \\
ContinualSkillBench~\citep{continualskillbench2026} & \cmark & \xmark & -- & -- & -- & -- & \xmark & \xmark & Cap. \\
RethinkSkill~\citep{rethinkskill2026} & \cmark & \cmark & \xmark & \cmark & \cmark & 3 & \cmark & \xmark & Cap. \\

\midrule
\rowcolor{highlightblue}
\textbf{RSE-Bench (Ours)} & \cmark & \cmark & \cmark & \cmark & \cmark & 4 & \cmark & \cmark & Cap. \\

\bottomrule
\end{tabular}
}
\vspace{-15pt}
\end{table*}


\section{Related works}

\textbf{Skill self-evolution in LLM agents.}
Agents can improve without weight updates by distilling execution experience into persistent artifacts: executable skill libraries \citep{voyager2023, skillweaver2025}, natural-language insights \citep{expel2024, reflexion2023}, workflow memory \citep{awm2024, suzgun2026dynamic}, and evolving playbooks \citep{ace2025}; recent methods add optimizer-style discipline with validation-gated skill edits \citep{skillopt2026, rethinkskill2026}. Unified codebases re-implement many such methods behind one interface for clean-condition comparison \citep{evomemory2026, memevolve2025}, and a wave of benchmarks measures whether evolution helps \citep{skilllearnbench2026, skillflow2026, continualskillbench2026, streambench2024}. All of these train and select skills from noise-free instructions, environments, and feedback; whether the evolution process itself survives realistic noise is left unexamined.


\textbf{Noise robustness of agents.}
LLMs degrade under irrelevant context \citep{gsmic2023}, and agents are similarly brittle to environmental distraction \citep{envdistraction2024, popup2024}. AgentNoiseBench \citep{agentnoisebench2026} injects controllable, solvability-preserving user- and tool-noise into agent benchmarks, and RethinkSkill \citep{rethinkskill2026} probes evolved skills with paraphrases and distractors. However, the former evaluates execution-time robustness of static agents, and the latter perturbs frozen skills only after evolution completes. As summarized in Table~\ref{tab:benchmark_comparison}, RSE-Bench extends evaluation to noise within the learning loop itself, where a noisy episode can be distilled into persistent skills and affect later tasks that reuse them.

\textbf{Contamination of agent memory and experience.} 
A security literature poisons agent persistence layers, including retrieval stores \citep{agentpoison2024, poisonedrag2024} and self-written memory \citep{minja2025, mpbench2026}. Recent work shows that self-evolution itself can be compromised: adversarial task exposure hardens into unsafe reusable skills \citep{skillmisevo2026}, and biased memory updates degrade agent safety \citep{memevobench2026}. These works inject deliberately malicious content and measure attack success, whereas we inject non-adversarial noise -- redundant instructions, irrelevant files, truncated logs, misbound records -- and measure capability on held-out tasks. Relatedly, reward misattribution across co-retrieved memories has been identified as an inherent flaw to mitigate \citep{romerl2026, attrimem2026}; our N-Bind instead treats outcome–record pairing as a controlled, injectable perturbation, measuring its isolated effect on skill evolution for the first time.

\section{RSE-Bench}

\subsection{Task Definition}
\label{sec:task_definition}

RSE-Bench evaluates \textbf{noise-robust skill self-evolution}: learning reliable, reusable skills from experience containing non-adversarial noise. Each noise condition targets one entry point in isolation (Section~\ref{sec:task_taxonomy}), while preserving task goals, required deliverables, ground-truth answers, native scoring, and per-task interaction budgets. Correct task evidence remains intact and accessible.

With the learned skill store frozen, we assess two capabilities on held-out tasks. \emph{Immunity} concerns performance on clean tasks after noisy learning. \emph{Adaptation} tests whether learning under input-side noise improves performance under the same noise compared with cleanly learned skills. Extended design rationale and the corresponding comparisons are provided in Appendix~\ref{app:design_principles}.

\subsection{Task Taxonomy}
\label{sec:task_taxonomy}

Every benchmark in RSE-Bench ships in five task conditions: one clean version and four noisy variants, each corrupting exactly one position of the learning loop under the task-preserving constraints of Section~\ref{sec:task_definition}.

\textbf{Instruction noise (N-Inst).}
Real users rarely hand an agent a minimal instruction: requests arrive wrapped in organizational background unrelated to the job, contextual details that do not affect the solution, and well-meant but unverified advice. N-Inst reproduces these habits with irrelevant filler that consumes attention, related-but-unnecessary information that invites false focus, and misleading non-binding suggestions, such as a hint to reuse a template that does not apply. The effective request is never altered: an agent that ignores the additions can solve the task exactly as before. N-Inst tests whether an agent treats unverified hints as binding constraints, and whether such hints survive distillation into its reusable rules.

\textbf{Evidence noise (N-Evid).}
Real workspaces are cluttered: directories accumulate outdated workbooks, retrieval returns reports from adjacent years, and near-duplicates abound. N-Evid injects such material into what the agent observes during execution: irrelevant files and pages, topically related but non-decisive evidence, near-miss artifacts that differ from the correct source in key fields, and redundant presentation such as repeated results. The correct evidence always remains reachable within the same interaction budget. N-Evid tests whether an agent can identify decisive sources amid clutter, and whether the selection habits it forms there carry into its skills.

\textbf{Record noise (N-Rec).}
Between execution and learning, logging failures can leave gaps in the recorded assistant text and tool outputs. N-Rec models such fragment loss by randomly selecting and deleting 32-character blocks from these fields, removing approximately 25\% of their characters. Only the record handed to the learner is perturbed; the execution itself, its outcome, and the environment state remain untouched. N-Rec tests whether a method can distill reliable skills from incomplete execution records and keep its reflections calibrated when the available evidence only partially captures how an outcome was reached.

\textbf{Binding noise (N-Bind).}
In deployed systems, learning is often asynchronous: outcomes are queued and batched, and delayed feedback can be attached to the wrong attempt. N-Bind pairs a task's outcome with another attempt's execution record from the same benchmark source and fit batch. Both the execution records and the outcomes are genuine; only their correspondence is altered. Applied only when the method itself decides to update, N-Bind tests whether a method can avoid drawing incorrect lessons from feedback attributed to the wrong attempt.

The four types differ in where they attach. N-Inst and N-Evid are materialized as data before a run begins, as noisy instructions and noisy environments; N-Rec and N-Bind operate at the boundary between execution and learning, and are applied through the harness at the moment the learner reads its record or consumes an update. N-Rec and N-Bind are easily conflated and deliberately separated: under N-Rec the record of this experience is altered, while under N-Bind the record is intact but belongs to a different experience. Both corrupt attribution, but through different mechanisms, and isolated injection lets us measure which one a given method resists better.

\subsection{Topic Categories}
\label{sec:topic_categories}

RSE-Bench spans four domains and six benchmarks: spreadsheet manipulation (SpreadsheetBench~\citep{spreadsheetbench2024}), document understanding (SP-DocVQA~\citep{spdocvqa2021} and MP-DocVQA~\citep{mpdocvqa2023}), interactive decision-making (ALFWorld~\citep{alfworld2021}), and sequential skill reuse (SkillFlow~\citep{skillflow2026} and ContinualSkillBench~\citep{continualskillbench2026}). These domains exercise procedural editing, evidence localization, multi-step planning, and transfer across ordered tasks, covering both textual and visual evidence. Detailed benchmark descriptions are provided in Appendix~\ref{app:benchmark_details}. All six benchmarks use deterministic native scoring rules shared by clean and noisy conditions, avoiding dependence on LLM judges whose interpretation might be affected by the injected content. The metrics and their application are detailed in Appendix~\ref{app:metrics}.

\subsection{Dataset Construction}
\label{sec:dataset_construction}

RSE-Bench varies noise along two axes. The first is \emph{where} noise enters, one of four entry points of the learning loop, perturbed one at a time; the second is \emph{which phase} it applies to, learning or held-out evaluation. Crossing the two (Table~\ref{tab:noise_cross}) separates what a single noisy run conflates: whether an agent can still execute in a degraded setting, and whether what it learned there is sound. All entry points share one clean-learning, clean-evaluation baseline, against which a degradation measured under clean evaluation is attributable to learning alone, since the exam never changed.

\textbf{Clean base.}
The clean base imports the original tasks, assets, and evaluators of the six sources introduced in Section~\ref{sec:topic_categories}, and supplies that baseline. Each source is split into an evolution-train partition on which self-evolution runs, a selection partition for method-internal decisions such as accepting a skill update, and a held-out final-test partition that is never learned from. Task identifiers are fixed, so noisy variants pair with clean counterparts, ensuring comparisons use identical tasks.

\textbf{N-Inst.}
Each item is its clean counterpart with text inserted into the user request, in one of three subtypes: content wholly outside the task's domain, a true but answer-irrelevant detail of the task's own material, and a pointer toward a different concern on that material that asserts nothing false. The two material-grounded subtypes are generated with the material in view, so the detail is real rather than invented. An insert may not change the goal, add binding constraints, or alter deliverables.

\textbf{N-Evid.}
N-Evid adds material to the space the agent searches during execution: records placed near the answer's evidence within a workbook or a page, extra pages in a multi-page task, or repeated genuine observations in an interactive environment. Generation is answer-aware: the generator receives the gold answer and the localized original evidence, because useful distractors live in a narrow margin, close enough to the question's topic to compete for attention yet never supplying a second correct answer or contradicting the real evidence, and knowing where the answer lives is what makes that margin checkable. The original evidence always remains intact and accessible, and distractors are introduced only as additions that never modify or obscure existing evidence. Benchmark-specific construction procedures and validity checks are detailed in Appendix~\ref{app:noise_details}.

\textbf{N-Rec.}
N-Rec perturbs the record of an episode after execution and scoring have completed, leaving both intact: the agent acted in a clean environment and was scored on what it did, and only the trace handed to the learner is incomplete. Two operators model loss and duplication under retry, deleting or repeating spans of assistant reasoning and tool output at a configurable rate. Deletion leaves no marker, filler, or cue, since a visible ellipsis would tell the learner exactly where to be cautious and thereby measure something easier. Evaluator feedback and citations are preserved verbatim, the episode's identity, status, and score are immutable, and so is the condition under which a method decides to update, so that trajectory noise cannot change how often learning happens.

\textbf{N-Bind.}
N-Bind models a failure of asynchronous learning pipelines in which feedback returning out of order is paired with execution records by arrival order rather than by identity, allowing an outcome to be paired with another attempt's record. Under this condition, every execution and evaluation remains genuine; the perturbation affects only their correspondence. For a misbound attempt, the learner receives the genuine evaluation result of another attempt from the same benchmark source and fit batch. Because delays are sampled randomly, the arrival order within a window may coincide with the original order, so not every episode is misbound. We record the actual number of misbound episodes in each batch and report the resulting misbinding rate in Section~\ref{sec:diagnostics}.

Table~\ref{tab:noise_cross} summarizes the conditions and the data behind them. N-Rec and N-Bind transform the experience consumed by the learner during skill updates, so these two run only as noisy learning under clean evaluation; they do not apply during held-out evaluation, when the skill store is frozen. The invariants above are enforced by automated checks where mechanically decidable, and otherwise by model review with human audits of sampled instances. The detailed automated verification pipeline, model-review prompts, and human audit procedures are provided in Appendix~\ref{app:noise_validation}.

\subsection{Data Statistics}
\label{sec:data_statistics}

Table~\ref{tab:data_statistics} summarizes the size and evolution-train, selection, and final-test splits of RSE-Bench. The two sequential benchmarks are split by stream position to preserve opportunities for later skill reuse. ALFWorld uses the official train, valid-seen, and valid-unseen splits, respectively, so its final-test environments are unseen during evolution.




\section{The \textsc{RSEClaw} Framework}
\label{sec:framework}

Two of the four entry points act on the experience a self-evolution method consumes during learning, so the conditions of Section~\ref{sec:dataset_construction} cannot be run from data alone; they require a framework that applies noise at consistent points in the evolution loop across methods. \textsc{RSEClaw} meets this requirement by decoupling the self-evolution method from the evolution loop: the method under evaluation is a module behind a fixed interface, while task delivery, native execution, scoring, and noise injection are managed by the framework. Methods implementing this interface can run in the same way under every condition without condition-specific modifications. Whether a particular noise type reaches a method, however, depends on the information that method actually consumes: for example, a method that does not read execution trajectories receives no perturbation under N-Rec. Such cases are explicitly recorded by the process diagnostics and are not treated as evidence of robustness.

\textbf{The evolution loop.}
\textsc{RSEClaw} follows the fixed cycle shown in Figure~\ref{fig:overview}. The framework delivers a task from the evolution-train partition, the method supplies whatever skills it judges relevant, and the framework executes the agent in the source's native environment and scores the outcome with the source's native evaluator; the resulting episode, comprising the instruction, the trajectory, and the evaluation, is then handed to the method, which may reflect on it, revise its skill store, or decline to update; the updated store is available from the next task onward. Selection tasks serve the method's internal decisions, and the held-out final-test partition is visited only after evolution ends, with the skill store frozen.

\textbf{The method interface.}
A method implements a small contract: retrieve skills relevant to an incoming task, consume a completed episode, decide whether to update, and expose its skill store for freezing and inspection. The contract deliberately excludes everything the experiment must control, so a method cannot re-score an episode, cannot reach the final-test partition, and cannot observe which condition it is running under.

\textbf{Noise seams.}
The two run-time conditions attach at seams the interface itself defines: N-Rec between execution and reflection, N-Bind where evaluations meet attempts. Their invariants are enforced at run time rather than by convention, so an implementation that mutates the original episode, alters an identity, status, or score, or shifts the method's update trigger is rejected at the point of violation instead of being discovered in analysis. The clean condition applies the identity transformation at the same seams, so clean and noisy runs traverse identical code paths.


\textbf{Process diagnostics.}
Final-test performance alone cannot distinguish a method that resisted noise from one that never encountered it. \textsc{RSEClaw} therefore records three types of process information at each optimization step. First, it records whether noise was applied and its realized extent: for $N_{\mathrm{Rec}}$, whether each episode was modified, the reason for any unmodified episode (e.g., ineligibility for learning or a method that does not consume trajectories), and the fraction of characters deleted; for $N_{\mathrm{Bind}}$, the number of mismatched episodes in each batch. Second, it records which episodes the method selected for learning. Third, it records whether a skill update was proposed, whether it passed validation and was accepted, and the proposed edits. The frozen skill store is supplied with every test task, so the deployed version determines which updates are available to the agent during evaluation. For methods that organise skills as individual rules, such as ACE, the framework also records the rules referenced by the agent on each test task. Whether a perturbation actually appears in the method's inputs is assessed offline. The framework logs every request received by the optimiser and provides paired-replay audits that compare clean and noisy optimiser inputs with all other conditions held fixed. These records help distinguish stable performance despite actual exposure and learning from cases in which noise never reached the learner or little learning occurred.

\section{Experiments}

\subsection{Evaluation Settings}
\label{sec:eval_settings}

We evaluate ACE~\citep{ace2025}, SkillOpt~\citep{skillopt2026}, and RethinkSkill~\citep{rethinkskill2026}, each implemented behind the \textsc{RSEClaw} interface of Section~\ref{sec:framework}, together with an otherwise identical base agent that maintains no skill store. The main experiments use GPT-5.2 and DeepSeek-V4.1-Flash; unless otherwise noted, analysis experiments use DeepSeek-V4.1-Flash. The evaluation protocol and detailed settings for the main experiments and analyses are provided in Appendix~\ref{app:setup}.



\begin{table*}[t]
\centering
\vspace{-15pt}
\caption{Main results on RSE-Bench (\%, means over three independent runs; standard deviations in Table~\ref{tab:run_variability}). Subscripts show relative changes, $100(s-b)/b$, from the corresponding $C\!\to\!C$ baseline $b$: green $\downarrow$ for decreases, gray $\uparrow$ for increases, and $\leftrightarrow$ for no change. Columns specify learning $\rightarrow$ evaluation; $C$ is clean, and Inst, Evid, Rec, and Bind denote instruction, evidence, record, and binding noise ($N_{\mathrm{Inst}}$--$N_{\mathrm{Bind}}$). Scores are macro-averaged equally over all six benchmarks before rates are computed. Underlines mark the largest relative decrease in each noise-condition column across the six method--model combinations, using unrounded rates and including exact ties. Per-benchmark results appear in Table~\ref{tab:main_results}.}
\label{tab:main_results1}
\begingroup
\fontsize{9}{10.4}\selectfont
\setlength{\tabcolsep}{1.6pt}
\renewcommand{\arraystretch}{1.12}
\newcommand{\rsebase}[1]{$#1$}
\newcommand{\rserate}[2]{%
  \begingroup
  \def\rsearrow{#1}%
  \def\rsedown{\downarrow}%
  \def\rseup{\uparrow}%
  \ifx\rsearrow\rsedown
    \color{green!50!black}%
  \else
    \ifx\rsearrow\rseup
      \color{black!55}%
    \else
      \color{black}%
    \fi
  \fi
  #1#2\%%
  \endgroup
}
\newcommand{\rsecell}[3]{%
  $#1_{\scalebox{0.78}{$\rserate{#2}{#3}$}}$%
}
\newcommand{\rseworst}[3]{%
  $\underline{#1}_{\scalebox{0.78}{$\rserate{#2}{#3}$}}$%
}
\resizebox{\textwidth}{!}{%
\begin{tabular}{@{}ll*{8}{c}@{}}
\toprule
\multirow{2}{*}{Method} & \multirow{2}{*}{Model}
& \multicolumn{2}{c}{Reference}
& \multicolumn{4}{c}{Noisy learning}
& \multicolumn{2}{c}{Noisy evaluation} \\
\cmidrule(lr){3-4}\cmidrule(lr){5-8}\cmidrule(l){9-10}
& & Initial & $C\!\to\!C$ & $N_{\mathrm{Inst}}\!\to\!C$ & $N_{\mathrm{Evid}}\!\to\!C$
& $N_{\mathrm{Rec}}\!\to\!C$ & $N_{\mathrm{Bind}}\!\to\!C$ & $C\!\to\!N_{\mathrm{Inst}}$ & $C\!\to\!N_{\mathrm{Evid}}$ \\
\midrule
\multicolumn{10}{@{}l}{\textbf{RSE-Bench (Avg)}} \\
\multirow{2}{*}{ACE} & DeepSeek-V4.1-Flash & \rsebase{71.25} & \rsebase{74.53} & \rsecell{72.64}{\downarrow}{2.5} & \rsecell{69.27}{\downarrow}{7.1} & \rsecell{73.08}{\downarrow}{1.9} & \rsecell{71.19}{\downarrow}{4.5} & \rseworst{69.88}{\downarrow}{6.2} & \rsecell{69.67}{\downarrow}{6.5} \\
 & GPT-5.2 & \rsebase{61.42} & \rsebase{63.97} & \rsecell{63.17}{\downarrow}{1.3} & \rsecell{64.01}{\uparrow}{0.1} & \rseworst{59.68}{\downarrow}{6.7} & \rsecell{61.54}{\downarrow}{3.8} & \rsecell{61.76}{\downarrow}{3.4} & \rseworst{58.65}{\downarrow}{8.3} \\
\multirow{2}{*}{RethinkSkill} & DeepSeek-V4.1-Flash & \rsebase{71.25} & \rsebase{75.93} & \rsecell{72.92}{\downarrow}{4.0} & \rsecell{71.31}{\downarrow}{6.1} & \rsecell{71.10}{\downarrow}{6.4} & \rsecell{74.32}{\downarrow}{2.1} & \rsecell{72.04}{\downarrow}{5.1} & \rsecell{71.92}{\downarrow}{5.3} \\
 & GPT-5.2 & \rsebase{61.42} & \rsebase{66.93} & \rseworst{62.93}{\downarrow}{6.0} & \rseworst{60.40}{\downarrow}{9.7} & \rsecell{64.74}{\downarrow}{3.3} & \rseworst{61.53}{\downarrow}{8.1} & \rsecell{63.57}{\downarrow}{5.0} & \rsecell{61.96}{\downarrow}{7.4} \\
\multirow{2}{*}{SkillOpt} & DeepSeek-V4.1-Flash & \rsebase{71.25} & \rsebase{77.71} & \rsecell{74.60}{\downarrow}{4.0} & \rsecell{75.24}{\downarrow}{3.2} & \rsecell{75.79}{\downarrow}{2.5} & \rsecell{76.20}{\downarrow}{1.9} & \rsecell{77.74}{\uparrow}{<0.1} & \rsecell{74.81}{\downarrow}{3.7} \\
 & GPT-5.2 & \rsebase{61.42} & \rsebase{67.53} & \rsecell{63.79}{\downarrow}{5.5} & \rsecell{65.06}{\downarrow}{3.7} & \rsecell{63.46}{\downarrow}{6.0} & \rsecell{63.38}{\downarrow}{6.2} & \rsecell{66.49}{\downarrow}{1.5} & \rsecell{63.29}{\downarrow}{6.3} \\
\bottomrule
\end{tabular}%
}
\endgroup
\vspace{-15pt}
\end{table*}

\subsection{Main Results}
\label{sec:main_results}

Table~\ref{tab:main_results1} reports final-test performance averaged over three independent runs of the complete evolution and evaluation pipeline; standard deviations are reported in Table~\ref{tab:run_variability}. The column headers specify the learning and evaluation conditions, with notation defined in the table caption. Under clean learning, self-evolution improves the six-benchmark macro-average for every method--backbone pair, with absolute improvements of $2.6$--$6.5$ percentage points over the initial agent.

\textbf{All four learning-side entry points erode the gain.} Of the $24$ noisy-learning cells in the Avg block, $23$ fall below their corresponding $C\!\to\!C$ baseline. Averaged over the six method--backbone pairs, relative losses are $4.9\%$ under $N_{\mathrm{Evid}}$, $4.5\%$ under $N_{\mathrm{Rec}}$, $4.4\%$ under $N_{\mathrm{Bind}}$, and $3.9\%$ under $N_{\mathrm{Inst}}$. Measured through differences in the overall macro-average score, these conditions retain only $25.5\%$--$42.1\%$ of the improvement achieved by clean evolution. $N_{\mathrm{Rec}}$ and $N_{\mathrm{Bind}}$ both cause larger average losses than $N_{\mathrm{Inst}}$, despite perturbing only the experience passed to the learner. How faithfully an agent's own execution is recorded thus matters as much as how clean the tasks it works on are, a comparison the input-side conditions alone cannot reveal.

\textbf{$N_{\mathrm{Evid}}$ is most damaging in both phases.} $N_{\mathrm{Evid}}$ causes a larger average loss than $N_{\mathrm{Inst}}$ whether injected during learning or evaluation. Its mean relative loss rises from $4.9\%$ during learning to $6.3\%$ during evaluation, whereas $N_{\mathrm{Inst}}$ causes losses of $3.9\%$ and $3.6\%$, respectively. The largest single decline occurs under evaluation-side $N_{\mathrm{Evid}}$: ACE with GPT-5.2 on SkillFlow falls from $31.33$ to $17.33$, a relative loss of $44.7\%$.

\textbf{Relative performance retention depends on the backbone.} SkillOpt achieves the highest clean macro-average on both backbones. On DeepSeek-V4.1-Flash, it also has the smallest mean relative loss across all six noisy conditions, combining a clean score of $77.71$ with a loss of $2.5\%$. On GPT-5.2, ACE has the smallest relative loss at $3.9\%$, alongside the lowest clean score of $63.97$. SkillOpt and RethinkSkill achieve higher clean scores of $67.53$ and $66.93$, but incur losses of $4.9\%$ and $6.6\%$, respectively. The clean-performance ranking is therefore consistent across backbones, while the method that best preserves its clean performance changes.

\textbf{Degradation varies across benchmarks.} Averaged over the $24$ noisy-learning cells within each benchmark, the largest losses occur on SkillFlow ($10.9\%$) and SpreadsheetBench ($5.7\%$), and the smallest on SP-DocVQA ($1.4\%$): a single-page extractive answer survives clutter largely intact, whereas a multi-step workbook edit depends on every cell the agent writes. ContinualSkillBench ($3.8\%$), ALFWorld ($3.4\%$), and MP-DocVQA ($3.1\%$) fall between. ContinualSkillBench also contains the largest positive deviations in the table, where SkillOpt improves by up to $23.2\%$ under noise, so an average over a block does not describe every cell within it. Finally, self-evolution does not improve every task even without noise: on GPT-5.2, ACE's clean evolution on SpreadsheetBench ends below the initial agent ($40.83$ vs.\ $45.00$).

\subsection{Process Diagnostics}
\label{sec:diagnostics}

Final-test performance alone cannot distinguish a method that resisted noise from one that never encountered it. We therefore examine three process quantities for ACE and RethinkSkill on SpreadsheetBench with DeepSeek-V4.1-Flash (Table~\ref{tab:noise_diagnostics}): input exposure rate (IER), the skill version selected on validation for final testing ($v^{*}$), and candidate-to-clean similarity (CCS). IER measures the share of optimiser inputs containing an injected perturbation across three rounds of $50$ fit tasks. CCS measures mean term-frequency cosine similarity between noisy-run and clean-run candidate skills. Its bracketed reference is computed over pairs of independent clean runs, and comparisons are made within each method.

\textbf{Actual exposure varies across methods and noise types.}
For ACE, IER is $100\%$ under $N_{\mathrm{Inst}}$ and $N_{\mathrm{Rec}}$, and $97\%$ under $N_{\mathrm{Evid}}$; for RethinkSkill, it is $100\%$ under $N_{\mathrm{Rec}}$. Binding noise affects a smaller share of inputs: $21\%$ ($32/150$) for ACE and $15\%$ ($22/150$) for RethinkSkill. These values measure actual mismatches between execution records and evaluation outcomes. Because random delays can preserve the original arrival order, some pairings remain unchanged even when binding noise is enabled. Within this two-method diagnostic subset, $N_{\mathrm{Bind}}$ also produces the smallest mean relative loss among the four learning-side conditions in Table~\ref{tab:main_results}.

\textbf{Different conditions select different skill versions.}
Under the same three-pass learning budget, the validation-selected version varies across conditions for both methods. ACE selects $v1$ under $N_{\mathrm{Rec}}$ and $v3$ under $N_{\mathrm{Evid}}$, while RethinkSkill selects $v1$ under $N_{\mathrm{Bind}}$ and $v3$ under $N_{\mathrm{Rec}}$. These differences identify which checkpoint from each learning trajectory is retained for final testing, providing additional information about the outcome of evolution.

\textbf{Lower candidate similarity can accompany a small change in final-test performance.}
All eight noisy-to-clean CCS values fall below their respective clean-to-clean reference means. The largest gap for ACE occurs under $N_{\mathrm{Evid}}$ ($0.702$ versus $0.814$), and for RethinkSkill under $N_{\mathrm{Bind}}$ ($0.783$ versus $0.901$). In the latter case, the mean final-test score changes from $88.33$ to $89.17$, a difference of only $0.84$ percentage points (Table~\ref{tab:main_results}); the corresponding standard deviations are $0.72$ and $3.15$ (Table~\ref{tab:run_variability}). This comparison illustrates that similar mean performance can coexist with different candidate text. CCS provides a complementary view of that textual variation, without directly establishing the functional quality of the selected skills.

\subsection{Analysis and Ablations}
\label{sec:ablation}


\textbf{Evaluation noise reverses the benefit of clean evolution.}
Table~\ref{tab:cross_noise_mini} and ~\ref{tab:cross_noise_spreadsheet} compare the initial agent with an agent carrying a cleanly evolved skill store. Under clean evaluation, evolution raises the score from $79.58$ to $90.00$. Under $N_{\mathrm{Inst}}$ evaluation, however, the initial agent scores $83.75$, while the cleanly evolved agent scores $77.50$; under $N_{\mathrm{Evid}}$, the corresponding scores are $82.50$ and $78.75$. The cleanly evolved agent therefore trails the initial agent by $6.25$ and $3.75$ percentage points on these noisy variants, reversing its advantage under clean evaluation.

\textbf{Additional rounds widen the clean--noisy gap.}
Appendix Table~\ref{tab:training_stages} compares checkpoints on clean held-out tasks after repeated passes over the same training set. From one to three rounds, the clean-learning score rises from $85.00$ to $90.00$, while the score under $N_2$ falls from $76.25$ to $73.75$. The clean--noisy gap widens monotonically from $8.75$ to $16.25$ percentage points.

\begin{wraptable}{r}{0.52\textwidth}
\centering
\captionsetup{skip=4pt}
\caption{Crossed noise conditions for ACE on SpreadsheetBench (DeepSeek-V4.1-Flash; scores in \%). Definitions and full results appear in Appendix Table~\ref{tab:cross_noise_spreadsheet}.}
\label{tab:cross_noise_mini}
\small
\setlength{\tabcolsep}{4pt}
\begin{tabular*}{\linewidth}{@{\extracolsep{\fill}}lccc@{}}
\toprule
& \multicolumn{3}{c}{Evaluation condition} \\
\cmidrule(lr){2-4}
Learning & Clean & $N_{\mathrm{Inst}}$ & $N_{\mathrm{Evid}}$ \\
\midrule
Initial               & 79.58 & 83.75 & 82.50 \\
Clean                 & 90.00 & 77.50 & 78.75 \\
$N_{\mathrm{Inst}}$    & 74.58 & 71.25 & --    \\
$N_{\mathrm{Evid}}$    & 74.17 & --    & 70.00 \\
\bottomrule
\end{tabular*}
\end{wraptable}

\textbf{Learning under matching noise performs worse than clean learning.}
The same table compares learning conditions while holding evaluation noise fixed. Under $N_{\mathrm{Inst}}$ evaluation, a store learned under $N_{\mathrm{Inst}}$ scores $71.25$, compared with $77.50$ for a cleanly learned store. Under $N_{\mathrm{Evid}}$ evaluation, learning under the matching noise scores $70.00$, compared with $78.75$ for clean learning. Matching the learning and evaluation noise thus reduces performance by $6.25$ and $8.75$ percentage points, respectively. For the method we test in this setting, exposure to the deployment noise during learning does not improve performance under that noise.



\textbf{GPT-5.5 reduces, but does not eliminate, noise-induced losses.}
For ACE on SpreadsheetBench, GPT-5.5 reduces the largest learning-side deficit from $15.83$ to $6.25$ percentage points compared with DeepSeek-V4.1-Flash (Appendix Table~\ref{tab:backbone_gpt55}). All four noisy-learning conditions nevertheless remain below its clean-learning score of $95.00$, with relative losses of $1.3\%$--$6.6\%$. These results show that learning-side degradation is not confined to the weaker backbone: a stronger model reduces its absolute magnitude in this setting but does not eliminate it.



\begin{figure*}[t]
    \centering
    \vspace{-20pt}
    \includegraphics[width=0.98\textwidth]{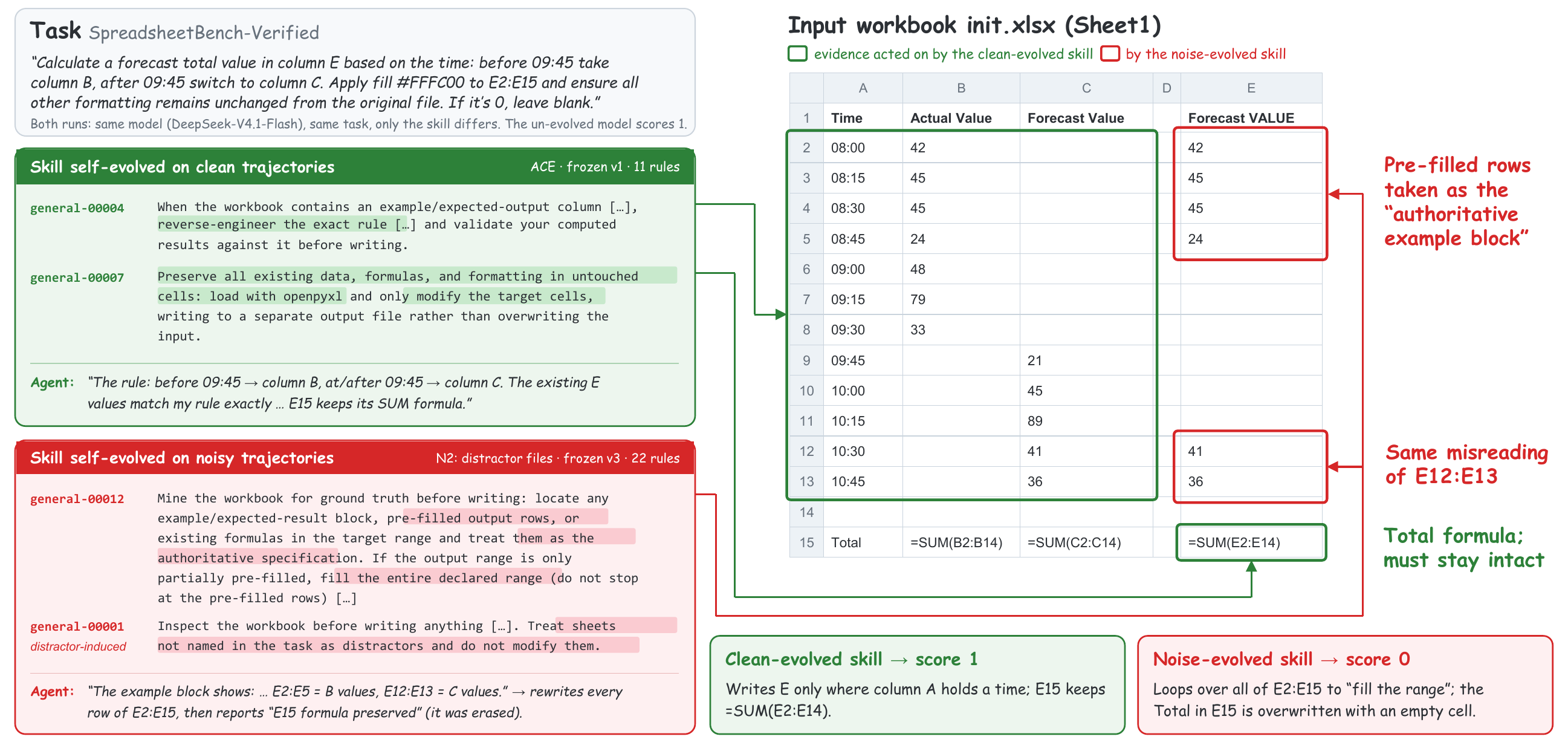}
    \caption{Case study: noisy evolution misreads pre-filled rows, overwrites the total formula, and flips the score from 1 to 0.}
    \label{fig:case_studies}
    \vspace{-20pt}
\end{figure*}



\label{sec:case_studies}
\textbf{Case Studies.}
Figure~\ref{fig:case_studies} illustrates one case, with detailed analyses in Appendix~\ref{app:case_studies}. These cases highlight two mechanisms: skills may retain assumptions or precautions without the conditions that justify them, making their reuse inappropriate; and reflection may combine useful content with unsupported corrections, including those inferred from misbound feedback. Validation assesses each candidate update as a whole, providing limited evidence about the contribution of individual clauses. Whether unsupported clauses persist in the deployed skill store therefore depends on reflection, update acceptance, and version selection; final-test scores alone cannot establish their presence.

\section{Conclusion}
\label{sec:conclusion}

RSE-Bench evaluates skill self-evolution under non-adversarial noise, separating learning-side from evaluation-side perturbations. Learning-side noise erodes a substantial fraction of the aggregate gain from clean evolution, with record and binding noise causing losses comparable to those from input-side noise. Candidate skill text can also diverge despite small changes in mean held-out scores, showing why performance alone cannot fully characterize learning under noise. We release RSE-Bench and \textsc{RSEClaw} to support joint evaluation of learning gains, noise exposure, and changes in learned skills.

\section*{AI Use Statement}
Large language models are the subject of study in this work: they serve as the evaluated actors, as the optimizers inside the ported self-evolution methods, and as the generator for the N-Inst and N-Evid noisy releases, all of which is described in the main text. Beyond these roles, we used LLM-based assistants to search for and locate related work, to draft and polish text under the authors' direction, and to help implement code; all research ideas, experimental designs, and claims are the authors' own, and the authors reviewed and take full responsibility for all text and code in this paper.

\section*{Ethics Statement}
RSE-Bench is constructed entirely from publicly available agent benchmarks used under their respective licenses; it involves no human subjects, no personally identifiable information, and no newly collected user data. The injected perturbations are non-adversarial by design: every noisy instance preserves the task goal, ground truth, scoring, and evidence reachability of its clean counterpart, so the benchmark measures robustness to naturally occurring noise rather than providing recipes for attacking deployed agents. We audited a sample of generated noisy instances to confirm they contain no harmful or offensive content. We see no ethical concerns beyond those inherent to evaluating LLM agents.

\section*{Reproducibility Statement}
We release the full benchmark and harness, including the five matched task releases per source benchmark, the noise-generation and review pipeline for N-Inst and N-Evid, and the runtime injection modules for N-Rec and N-Bind, which are deterministic given a recorded seed and make no model calls. Every run records its resolved configuration, per-phase noise conditions, model identifiers, budgets, and frozen checkpoint, so any reported number can be regenerated from the released artifacts.

\bibliography{iclr2027_conference}
\bibliographystyle{plainnat}

\clearpage
\appendix
\makeatletter
\def\ttl@Hy@refstepcounter#1{%
  \let\rse@saved@hyper@refstepcounter\hyper@refstepcounter
  \def\hyper@refstepcounter##1{%
    \hyper@makecurrent{##1}%
    \edef\ttl@Hy@saveanchor{%
      \noexpand\Hy@raisedlink{%
        \noexpand\hyper@anchorstart{\@currentHref}%
        \noexpand\hyper@anchorend}}%
  }%
  \refstepcounter{#1}%
  \let\hyper@refstepcounter\rse@saved@hyper@refstepcounter
}
\patchcmd{\ttl@select}{\ttl@Hy@saveanchor}{}{}%
  {\PackageError{appendix-toc}{Cannot defer heading anchors}{Check the titlesec version.}}
\patchcmd{\ttl@select}
  {\everypar{\setbox\z@\lastbox\ttl@strut}}
  {\everypar{\setbox\z@\lastbox\ttl@Hy@saveanchor
    \let\ttl@Hy@saveanchor\@empty\ttl@strut}}{}%
  {\PackageError{appendix-toc}{Cannot attach heading anchors}{Check the titlesec version.}}
\makeatother
\addtocontents{toc}{\protect\setcounter{tocdepth}{2}}
\begingroup
  \hypersetup{linkcolor=jingdongred,linktoc=section}
  \section*{Supplementary Material}
  \subsection*{Contents}
  \makeatletter
  \renewcommand{\numberline}[1]{#1\quad}
  \renewcommand{\l@section}[2]{%
    \ifnum\c@tocdepth>\z@
      \vskip6pt\noindent#1\dotfill#2\par
    \fi}
  \renewcommand{\l@subsection}[2]{%
    \ifnum\c@tocdepth>\@ne
      \vskip6pt\noindent\hspace*{1em}#1\dotfill#2\par
    \fi}
  \@starttoc{toc}
  \makeatother
\endgroup
\par\kern20pt

\section{Design Principles and Evaluation Objectives}
\label{app:design_principles}

\textbf{Noise characterized by entry point.}
RSE-Bench distinguishes noise by where it enters the learning loop: task instructions, environmental evidence, the execution record consumed by the learner, or the pairing between a record and its evaluation outcome. These positions represent different failure modes, including distracting context, cluttered workspaces, missing log fragments, and misattributed feedback. They are not levels on a common severity scale. Each condition perturbs one entry point in isolation, supporting controlled comparisons of how methods respond to different sources of unreliable experience.

\textbf{Task-preserving comparisons.}
Clean and noisy conditions retain the same task goals, required deliverables, ground-truth answers, and native evaluators. They also share the same per-task interaction budgets, including the step, time, and output limits of Appendix~\ref{app:setup_agent}. The correct evidence remains reachable. Noise can increase the difficulty of finding relevant evidence or interpreting experience without changing what counts as success. N-Rec and N-Bind additionally preserve the underlying executions and genuine evaluations, modifying only the records delivered to the learner or their correspondence with outcomes.

\textbf{Consequences beyond the learning episode.}
A local misjudgment may affect later tasks if it is incorporated into a reusable skill. Evaluating only the episode in which noise appears can therefore miss its downstream consequences. RSE-Bench assesses the resulting skill store on held-out tasks after learning has ended, distinguishing two objectives.

\emph{Immunity} is assessed by comparing noisy learning followed by clean evaluation ($N\!\to\!C$) with clean learning and clean evaluation ($C\!\to\!C$). This comparison measures how well skills learned from noisy experience retain their usefulness on clean tasks.

\emph{Adaptation} is assessed by comparing matched noisy learning and evaluation ($N\!\to\!N$) with clean learning under the same noisy evaluation ($C\!\to\!N$). This tests whether exposure during learning improves performance under that noise. The comparison applies to N-Inst and N-Evid; N-Rec and N-Bind operate only during learning and therefore have no noisy-evaluation counterpart.

\section{Benchmark Details}
\label{app:benchmark_details}

RSE-Bench combines six benchmarks across four domains to evaluate skills that differ in both their content and their conditions of reuse. Evidence ranges from structured workbooks and document images to interactive environment states, while ordered task streams provide opportunities to examine transfer from earlier learning to later tasks.

\textbf{Spreadsheet manipulation.}
The agent modifies real workbooks according to natural-language requests, including filling formulas, adjusting formats, merging records, and maintaining cross-sheet references. Useful skills are procedural, such as inspecting a worksheet's layout before editing or checking referenced ranges after modifying a formula. Such procedures can be reused across workbooks, making it important to distinguish broadly applicable rules from assumptions specific to one file. SpreadsheetBench~\citep{spreadsheetbench2024} provides tasks drawn from real user requests, with outputs checked against ground-truth workbooks at the cell level.

\textbf{Document understanding.}
The agent answers questions grounded in document images, requiring it to locate relevant pages and regions and interpret forms, tables, and layout. Skills concern evidence handling: where to search, what to cross-check, and when enough evidence has been found. SP-DocVQA~\citep{spdocvqa2021} poses questions over single document images, whose answers depend on within-page content and structure. MP-DocVQA~\citep{mpdocvqa2023} extends this setting to multi-page documents, requiring answer-page localization in addition to reading the relevant content. Together, these benchmarks evaluate evidence-handling skills across within-page reading and cross-page search.

\textbf{Interactive decision-making.}
The agent acts over multiple steps, with success determined by its actions and the resulting environment state. ALFWorld~\citep{alfworld2021} provides household tasks in a text environment, exercising planning over object locations and action sequences. Reusable skills include behavioral routines such as deciding where to search for an object, ordering actions to satisfy prerequisites, and verifying that a subgoal has been completed.

\textbf{Sequential skill reuse.}
Ordered task streams provide repeated opportunities to apply skills acquired from earlier experience. SkillFlow~\citep{skillflow2026} groups tasks into families with shared workflow structures while varying entities, file formats, and difficulty. Within each family, tasks follow a fixed order, supporting the reuse of procedures distilled from earlier tasks. ContinualSkillBench~\citep{continualskillbench2026} provides longer streams of subtasks ordered by skill-transfer dependencies and increasing difficulty, one stream per domain (finance and economics, healthcare, law, and mathematics), so that skills acquired on earlier subtasks can support later ones that build on them. This structure supports analyses of how earlier learning relates to later performance; final-test evaluation in RSE-Bench still follows the shared protocol with the skill store frozen.

\textbf{Scoring.}
Every benchmark is scored by its native evaluator: cell-level comparison against a ground-truth workbook for SpreadsheetBench, exact match against the gold answers for SP-DocVQA and MP-DocVQA, the simulator's goal-satisfaction flag for ALFWorld, the task's test suite for SkillFlow, and test suites, exact match, numerical tolerance, or token-level F1, depending on the task, for ContinualSkillBench. Clean and noisy conditions retain the same ground truth and scoring rules. No language model is involved in scoring, which keeps evaluation deterministic and avoids an additional source of variation from LLM-based judging. Benchmark-specific metric definitions are provided in Appendix~\ref{app:metrics}.

\section{Metrics}
\label{app:metrics}

Every benchmark is scored by its own native evaluator, called through the harness without altering its answer rules, and exactly one metric per benchmark is reported. Table~\ref{tab:metrics} lists them. Noise never reaches the evaluator: a noisy variant of a task carries the same gold answer, golden workbook, answer position and test suite as its clean counterpart --- on SpreadsheetBench, for instance, the target records of the $N_{\mathrm{Inst}}$ and $N_{\mathrm{Evid}}$ variants are byte-identical to the clean ones for all $160$ tasks of the pool --- so a clean and a noisy run of the same task are graded by the same fixed rule.

\begin{table}[t]
\centering
\caption{Reported metric per benchmark. \emph{Soft} is a partial-credit signal that the harness exposes to a method's own selection gate but never reports as a result; a benchmark without genuine partial credit reports none rather than a copy of the binary outcome.}
\label{tab:metrics}
\vspace{6pt}
\small
\setlength{\tabcolsep}{5pt}
\resizebox{\textwidth}{!}{%
\begin{tabular}{@{}llll@{}}
\toprule
Benchmark & Reported metric & What it checks & Soft \\
\midrule
SpreadsheetBench      & \texttt{hard\_restriction} $\in \{0,1\}$ & cell equality over the declared answer ranges & --- \\
SP-DocVQA             & \texttt{accuracy} $\in \{0,1\}$         & match against any gold answer string          & ANLS \\
MP-DocVQA             & \texttt{accuracy} $\in \{0,1\}$         & match against any gold answer string          & ANLS \\
ALFWorld              & \texttt{won} $\in \{0,1\}$              & goal satisfaction in the simulator            & --- \\
SkillFlow             & \texttt{reward} $\in \{0,1\}$           & exit code of the task's native test suite     & --- \\
ContinualSkillBench   & \texttt{reward} $\in [0,1]$             & native judge: tests, match, tolerance, or F1  & F1 subset \\
\bottomrule
\end{tabular}%
}
\end{table}

\textbf{SpreadsheetBench.} The agent submits a path to a workbook. The harness opens it with \texttt{openpyxl} to reject unreadable files, has LibreOffice recalculate it so that submitted formulas are compared by their values, and then calls the native cell-level comparison against the golden workbook over the answer positions the task declares, which may be several \texttt{sheet!range} fragments. Values are normalised before comparison: numbers are rounded to two decimals, numeric strings are coerced to numbers, datetimes become Excel serial days rounded to integers, and times are truncated to minutes; an empty cell and an absent cell count as equal. A task scores $1$ only if every cell of every declared range matches, so the metric is all-or-nothing per task, and cell formatting is not compared. The companion \texttt{soft\_restriction} field is the same boolean cast to a float in this harness rather than a partial score, and is therefore not used as one.

\textbf{SP-DocVQA and MP-DocVQA.} Both call the same vendored DocVQA evaluator on the submitted answer string. Reported \texttt{accuracy} is $1$ if the answer, lowercased and stripped, equals any one of the gold answers, and $0$ otherwise. The evaluator also returns ANLS, one minus the normalised Levenshtein distance to the closest gold answer, floored to $0$ below a similarity of $0.5$; it is retained as the soft signal and not reported. The two benchmarks differ only in what the agent must do to obtain the answer, not in how the answer is graded: MP-DocVQA supplies multiple pages, so page localisation is part of the task rather than a separately scored quantity.

\textbf{ALFWorld.} The episode runs against the native TextWorld PDDL environment, and \texttt{won} is the environment's own goal-satisfaction flag when the episode ends. The environment is constructed deterministically --- fixed seed, no shuffling of admissible commands --- so the same action sequence yields the same verdict. ALFWorld defines no partial credit, and no reward shaping is added.

\textbf{SkillFlow.} After execution, the harness copies the task's native test files into the deployed workspace and runs \texttt{pytest} in the task's working directory; \texttt{reward} is $1$ if the suite exits zero and $0$ otherwise. Grading is tied to the workspace the harness deployed rather than to whatever path the agent names in its submission, so an agent that did the work but answered in prose is still graded on its files. A per-test pass fraction could be reconstructed from the test report, but it would be our construct rather than the benchmark's, so it is neither used nor reported.

\textbf{ContinualSkillBench.} The native judge grades the JSON output file the task asked for, and \texttt{reward} is the score it returns. The released pool of tasks declares four evaluation types and no rubric: one class of programmatic tasks scored $1.0/0.0$ by their \texttt{pytest} suite; one class of exact-match tasks compared after lowercasing, mapping punctuation to spaces and collapsing whitespace; one class of numeric tasks compared under a per-task relative tolerance (default $10^{-4}$), falling back to the string comparison when either side does not parse as a number; and one class of F1 tasks scored by token-level F1 after lowercasing, stripping punctuation and dropping articles. The F1 subset is the only place where a reported score is continuous rather than binary. Because no task in the pool carries rubrics, the judge's model-scored branch is never entered, and it is the only branch in any of the six benchmarks that would involve a language model in scoring: no LLM judge participates in any number reported in this paper.

\textbf{Aggregation.} Per-benchmark scores are means of the reported metric over the tasks of the split, in percent. The macro-average over benchmarks weights the six equally and is computed before any relative change is taken, so a rate in Table~\ref{tab:main_results1} is the change of the macro-average and not an average of per-benchmark rates.

\section{Experimental Setup}
\label{app:setup}

This appendix specifies the protocol behind every number in Section~\ref{sec:main_results} and Section~\ref{sec:ablation}. Configuration values are those of the released run configurations; where a value differs between benchmarks, the difference is stated.

\subsection{Evaluation Protocol}
\label{app:setup_protocol}

\textbf{Jobs.}
A job is one benchmark, one method, one backbone and one learning condition. Every job starts from the method's own initial skill store (empty for RethinkSkill and SkillOpt, an empty two-section playbook for ACE) and shares no state with any other job, so the five learning conditions of a method--backbone--benchmark triple are five independent evolutions over the same tasks.

\textbf{Partitions.}
Each benchmark contributes three disjoint partitions (Table~\ref{tab:data_statistics}). The $50$ evolution-train tasks are drawn from the benchmark's native training split and the $30$ selection tasks from its native validation split, which is never merged into training; both are drawn once with a fixed seed ($42$) and whole groups of tasks that share material are kept together. The final-test partition is drawn from the native test split. The evolution-train and selection identifiers are fixed before any run. All identifiers are identical across methods, backbones and conditions, so every comparison in the paper is paired at the task level.

\textbf{The evolution run.}
A run makes $R=3$ passes over the evolution-train partition. Before the first pass the initial store is scored on the selection partition. During a pass the agent executes each training task with the current store, the episodes are handed to the method, and the method proposes an update at the points its own algorithm defines (once per pass for ACE and RethinkSkill, once per batch for SkillOpt; Appendix~\ref{app:setup_methods}). Every proposed store is scored on the selection partition, and the method's own gate decides whether it replaces the current one. After the last pass the harness freezes, among the checkpoints the method promoted, the one with the highest selection score, breaking ties toward the earlier version; if no checkpoint is comparable, the initial store is frozen. The frozen store is then evaluated once on the final-test partition. No method sees a final-test task before this point, and the store no longer changes during evaluation.

\textbf{Learning and evaluation conditions.}
The column headers of Table~\ref{tab:main_results1} name a learning condition and an evaluation condition.
\begin{itemize}
\item \emph{Clean learning} ($C\!\to\!\cdot$) uses the clean evolution-train and selection partitions.
\item \emph{$N_{\mathrm{Inst}}$ and $N_{\mathrm{Evid}}$ learning} replace both the evolution-train and the selection partition with the corresponding noisy release, task for task, so the method's gate and the frozen-version choice also operate on noisy tasks. The noisy releases carry the same gold answers and evaluators as the clean ones (Appendix~\ref{app:metrics}).
\item \emph{$N_{\mathrm{Rec}}$ and $N_{\mathrm{Bind}}$ learning} use the clean partitions and apply the transformation at the run-time seams of Section~\ref{sec:framework}, to training episodes only. Selection episodes and final-test episodes are never transformed.
\item \emph{Clean evaluation} ($\cdot\!\to\!C$) runs the frozen store on the clean final-test partition. \emph{Noisy evaluation} ($C\!\to\!N_{\mathrm{Inst}}$, $C\!\to\!N_{\mathrm{Evid}}$) reloads the store frozen by the clean-learning job and runs it on the corresponding noisy final-test partition, without any further learning.
\end{itemize}
The \emph{initial agent} is the same actor with no skill store. For each benchmark--backbone pair, it is evaluated once per independent
run on the clean final-test partition, and the resulting score is
shared by the three methods within that run. For the noisy-evaluation
analyses, it is also evaluated once per run on each noisy final-test
partition ($\mathrm{Initial} \rightarrow N$).

\textbf{Runs.}
Each cell of Table~\ref{tab:main_results1} is the mean over 3 independent runs of the complete pipeline, evolution and evaluation included; standard deviations are in Table~\ref{tab:run_variability}.

\subsection{Agent, Backbones and Budgets}
\label{app:setup_agent}

\textbf{Actor.}
All methods share one tool-calling actor, which runs in the benchmark's native environment through the local executor. It has a fixed system instruction (``Complete the given benchmark task using tools. Files use real local paths and persist across tool calls. Finish by calling submit with the specified submission format.''), receives the task observation as JSON together with any page images, and acts only through tool calls: \texttt{python} and \texttt{command} execute code in the task workspace, \texttt{read\_file} and \texttt{write\_file} access workspace files, \texttt{env\_step} sends an action to an interactive environment (ALFWorld only), and \texttt{submit} ends the episode with the benchmark's native submission object. A reply without a tool call, or with arguments that are not valid JSON, receives a one-line reminder of the protocol. The skill store is inserted as a separate system message after the fixed instruction and before the task, so it is never mistaken for part of the question; it carries no version number, so the actor cannot tell how far learning has progressed. ACE additionally asks the actor to cite the playbook bullets it relies on, as in the original method.

\textbf{Per-task limits.}
Each episode is limited to $40$ tool steps, $60$\,s per tool call and $600$\,s of wall-clock time. The standard output and error of each tool call are truncated to $16{,}000$ characters, keeping head and tail. Because every step resends the transcript, the three most recent tool results are kept whole and older ones are shortened to $500$ characters by the same head-and-tail rule; the actor can re-run a call to recover an old result. The current step's input is never shortened.

\textbf{Backbones and roles.}
Table~\ref{tab:setup_models} lists the model settings. Within a job the same backbone serves as the actor and as the method's optimizer (Reflector and Curator for ACE, analysts, merger, ranker and epoch updater for SkillOpt, rewriter for RethinkSkill), so a backbone column of Table~\ref{tab:main_results1} describes a self-contained system. The $N_{\mathrm{Inst}}$ and $N_{\mathrm{Evid}}$ variants (Appendix~\ref{app:noise_validation}) are constructed and reviewed offline by Kimi-K2.8, a model outside the evaluated backbones that never participates in an evolution run. The generator and both semantic reviewers use this model with the settings in the lower block of Table~\ref{tab:setup_models}; the two reviews are separate requests that do not see each other's verdicts. Decoding is deterministic, and every candidate, including rejected ones, is kept with its review records. No language model takes part in scoring (Appendix~\ref{app:metrics}).

\begin{table}[t]
\centering
\caption{Model settings. Upper block: evaluated backbones; the actor and the optimizer of a job use the same backbone and the same settings. Lower block: the offline noise-construction model, used for generation and review only. Maximum output and timeout are per request; all models are called through the Chat Completions protocol. For GPT backbones the temperature is left at the provider default.}
\label{tab:setup_models}
\vspace{6pt}
\small
\setlength{\tabcolsep}{5pt}
\resizebox{\textwidth}{!}{%
\begin{tabular}{@{}lllccc@{}}
\toprule
Model & Role & Temperature & Max output (tokens) & Timeout (s) & Thinking mode \\
\midrule
DeepSeek-V4.1-Flash & actor, optimizer (main, analyses)    & $0$     & $32{,}768$ & $600$ & disabled \\
GPT-5.2             & actor, optimizer (main)              & default & $32{,}768$ & $600$ & provider default \\
GPT-5.5             & actor, optimizer (backbone analysis) & default & $32{,}768$ & $600$ & provider default \\
\midrule
Kimi-K2.8           & $N_{\mathrm{Inst}}$/$N_{\mathrm{Evid}}$ generator, reviewer & $0$ & $8{,}192$ & $180$ & disabled \\
\bottomrule
\end{tabular}%
}
\end{table}

\subsection{Method Configurations}
\label{app:setup_methods}

The three methods are reimplemented against their published source, keeping each method's update schedule, evidence, prompts and acceptance rule; only rollout, scoring and noise injection are delegated to the framework. Table~\ref{tab:setup_methods} summarizes the differences that matter for noise exposure.

\textbf{ACE.}
After each pass, a Reflector call is made for every training episode, successes included. It receives the question, the recorded trajectory truncated to $6{,}000$ characters, the submission, the evaluator feedback and the playbook bullets the actor cited, and returns a reflection with helpful/harmful/neutral tags for those bullets. The tags update the bullets' counters, and a single Curator call then turns the reflections into operations on the playbook. As in the published code, only \texttt{ADD} operations are applied, the optional deduplicating analyzer is disabled, and there is no score-based acceptance gate: every proposed playbook replaces the current one. The playbook starts with two empty sections, \emph{general} and \emph{others}, so that the published rule redirecting unknown sections to \emph{others} is meaningful from the first pass. The selection partition therefore affects ACE only through the choice of the frozen version.

\textbf{SkillOpt.}
SkillOpt updates per step: the $50$ training tasks of a pass are split into steps of $40$ and $10$ tasks, giving six steps over three passes. In each step, analyst calls read the trajectories of failed and of successful episodes in minibatches of $8$ and each proposes at most $L$ edits, the proposals are merged hierarchically in groups of $8$, and a ranking call keeps the top $L$. The edit budget $L$ follows a cosine schedule from $4$ down to $2$ over the run. The patched skill is accepted only if its selection score strictly exceeds that of the current skill, so a tie is rejected. From the second pass on, an epoch-end update runs the previous and the current epoch's skills on the same $20$ training tasks, sampled with a fixed seed, and uses the paired outcomes to rewrite a reserved slow-update section of the skill (always accepted, as in the published default) and a meta-skill that conditions the optimizer's later prompts without being shown to the actor. On SP-DocVQA, the gold answers and ANLS of scored training episodes are appended to the learning record after evaluation, reproducing the published DocVQA rollout; the actor never sees them and held-out episodes are not affected.

\textbf{RethinkSkill.}
RethinkSkill updates once per pass. Its evidence is the evaluation ledger of all training episodes: per task a success or failure label (success only at a hard score of exactly $1$), the hard score, the soft score where the benchmark has one, and the evaluator's own reason string where it emits one. A single optimizer call returns the complete replacement skill as a JSON object with exactly three string fields (\texttt{operation}, \texttt{candidate\_skill}, \texttt{rationale}); any other reply, and an explicit \texttt{noop}, leaves the skill unchanged. The candidate is judged by a four-branch gate on the selection partition: it is accepted if the hard score rises by more than a dead band, rejected if it falls by more than the band, accepted as a soft rescue if the hard score is within the band but the soft score rises by at least a rescue margin, and otherwise left flat. The dead band and rescue margin are the published per-benchmark values: $0.01$ and $0.02$ for SP-DocVQA and MP-DocVQA, $0.02$ and $0.05$ for SpreadsheetBench, $0.02$ with no rescue for ALFWorld. The published protocol does not cover SkillFlow and ContinualSkillBench, so a band of $0$ with no rescue is used there instead of an invented value.

\begin{table}[t]
\centering
\caption{Method configurations. \emph{Reads trajectory} determines whether $N_{\mathrm{Rec}}$ has a route into the method's learning input; all three methods receive $N_{\mathrm{Inst}}$, $N_{\mathrm{Evid}}$ and $N_{\mathrm{Bind}}$ through their episodes or outcomes.}
\label{tab:setup_methods}
\vspace{6pt}
\small
\setlength{\tabcolsep}{4pt}
\resizebox{\textwidth}{!}{%
\begin{tabular}{@{}lllll@{}}
\toprule
Method & Update point & Learning input & Reads trajectory & Acceptance \\
\midrule
ACE          & once per pass       & every training episode, per-episode reflection & yes & none (always accept) \\
SkillOpt     & per step (40 / 10) + epoch end & failed and successful trajectories, minibatch 8 & yes & strict improvement on selection \\
RethinkSkill & once per pass       & evaluation ledger of every training episode   & yes/no  & four-branch gate on selection \\
\bottomrule
\end{tabular}%
}
\end{table}

\subsection{Run-Time Noise Parameters}
\label{app:setup_noise}

\textbf{$N_{\mathrm{Rec}}$.}
The main experiments use the deletion operator at severity $0.25$ on two channels, the assistant's text and the tool outputs, with a fixed seed ($42$). Evaluator feedback and bullet citations are preserved, no marker is left at a deletion, and at most $2{,}000$ characters may be added by any operator. The transformation is applied to training episodes, including the paired episodes of SkillOpt's epoch-end update, and never to selection or final-test episodes.

\textbf{$N_{\mathrm{Bind}}$.}
Outcomes are bound to execution records by arrival order within windows of $3$ attempts, with delays drawn uniformly up to $6$ positions from a fixed seed ($42$). Binding is applied within each training batch (a pass for ACE and RethinkSkill, a step for SkillOpt) and, for SkillOpt, separately to the previous-epoch and current-epoch sides of the epoch-end comparison, never across them. Where the sampled arrival order within a window coincides with the dispatch order, no outcome moves; the binding record of every batch counts the episodes whose outcome was actually reordered.

\subsection{Analysis Experiments}
\label{app:setup_analysis}

All analyses use ACE on SpreadsheetBench with DeepSeek-V4.1-Flash, the same partitions and the same settings as the main experiments, except where noted below.

\textbf{Crossed learning and evaluation} (Table~\ref{tab:cross_noise_spreadsheet}). The initial agent is evaluated on the clean and on both noisy final-test partitions. Stores frozen by clean learning are evaluated on the noisy partitions ($C\!\to\!N$), and stores frozen by $N_{\mathrm{Inst}}$ and $N_{\mathrm{Evid}}$ learning are evaluated on the clean partition ($N\!\to\!C$) and on the noisy partition of the same type ($N\!\to\!N$). All of these reuse frozen stores; no store is re-learned for the analysis.

\textbf{Training rounds} (Table~\ref{tab:training_stages}). This ablation uses one clean-learning run and one \(N_2\)-learning run, each making three passes over the same evolution-train partition. For each condition, we evaluate the skill checkpoint at the end of each round \(R \in \{1,2,3\}\) on the clean final-test partition, with the skill store frozen during evaluation. The three columns thus track round-end checkpoints along a single evolution trajectory within each condition, including the checkpoint at the end of the third round for \(R=3\). The gap at each round is the noisy-learning score minus the clean-learning score, in percentage points.

\textbf{Backbone} (Table~\ref{tab:backbone_gpt55}). The initial agent, clean learning, the four noisy-learning conditions and the two noisy-evaluation conditions are repeated with GPT-5.5 as both actor and optimizer, with the settings of Table~\ref{tab:setup_models}.

\textbf{Process diagnostics} (Section~\ref{sec:diagnostics}). IER, the selected version $v^{*}$ and CCS are computed offline from the logged learning inputs, gate records and candidate skills of the main-experiment runs of ACE and RethinkSkill; they require no additional model calls.

\textbf{Optimizer-request logs and paired replay.}
\label{app:setup_audit}
Every optimizer call is recorded as it is made: the full prompt, the effective system message, the request arguments and whether the call returned or failed. These logs are passive and change nothing the method sends or receives. They let the analyses check what a method's optimizer actually received, rather than inferring it from what the noise seam produced; for example, an episode modified by $N_{\mathrm{Rec}}$ that a method never selects, or that a method reads only through its evaluation ledger, does not reach the optimizer. To confirm that the run-time conditions behave as intended, we additionally run controlled paired replays offline: the same method state, task order and scripted optimizer replies are replayed once with the identity transformation and once with $N_{\mathrm{Rec}}$ or $N_{\mathrm{Bind}}$, and the corresponding requests are compared for exact prompt equality. A comparison is counted only when the system message and request arguments are unchanged and the number of calls matches; under $N_{\mathrm{Bind}}$, whose reordering can change which episodes a method selects, requests are paired by the receiving episode before comparison. The replays use no model and no additional evolution runs; they serve as a check of the harness and of the logs, not as a separate experiment.

\section{N-Evid Construction Details}
\label{app:noise_details}

This appendix details how the N-Evid releases are built for each benchmark. Three construction families cover the six sources, chosen by what the agent actually searches during execution: rendered page images for the two DocVQA sources, input files in a task workspace for SpreadsheetBench, SkillFlow and ContinualSkillBench, and the stream of simulator observations for ALFWorld. Table~\ref{tab:nevid_routes} summarizes the construction family and gold-answer access for each benchmark.

\textbf{Shared protocol.}
All families share four rules. (i) \emph{Group granularity.} Tasks that share the same native material (the same page image, workbook or environment layout) form a group, and one piece of noise material is constructed for the whole group, since the agent observes the same material whichever of the group's questions it answers. Every question of the group is checked separately, and the group is published only if all of them pass; a single rejection withdraws the whole group rather than a single task. (ii) \emph{Additions only.} The native question, gold answers, evaluator and train/test assignment are copied unchanged, and every original asset remains present and byte-identical wherever the construction does not by design append to it; this is verified by mechanically replaying each renderer over the untouched original. (iii) \emph{Separated roles.} Generation and review are separate model calls at temperature $0$; reviewers receive the final rendered material rather than the generator's rationale, and a verdict of \emph{uncertain}, a malformed verdict or a transport failure counts as non-acceptance. (iv) \emph{Bounded and recorded.} Each group has a fixed ceiling on logical model calls, all requests, responses, rejections and failures are retained, and resuming a run never re-draws a rejected group.

\begin{table}[t]
\centering
\caption{N-Evid construction families and gold-answer access by benchmark. Relation labels are assigned per group by a seeded hash; the workspace-reference fallback is always related non-decisive.}
\label{tab:nevid_routes}
\vspace{6pt}
\small
\setlength{\tabcolsep}{5pt}
\resizebox{\textwidth}{!}{%
\begin{tabular}{@{}lll@{}}
\toprule
Benchmark & Family & Generator sees gold \\
\midrule
SP-DocVQA           & answer-aware page edit & yes \\
MP-DocVQA           & answer-aware page edit & yes \\
SpreadsheetBench    & workspace evidence     & no (hidden) \\
SkillFlow           & workspace evidence     & no (hidden) \\
ContinualSkillBench & workspace evidence     & no (hidden) \\
ALFWorld            & observation replay     & not applicable \\
\bottomrule
\end{tabular}%
}
\end{table}

\subsection{Answer-Aware Page Edits: SP-DocVQA and MP-DocVQA}

\textbf{Localization.}
The builder first asks the model to localize the evidence of every question in the group on the original page images, given the questions and their gold answers. Each piece of evidence is returned as a bounding box in original pixel coordinates together with the text observed inside it, and must include field labels and headers needed to read the value; the model also lists the \emph{qualifiers}, the conditions and document identity that select the answer (e.g.\ a year, an entity, a form section). Boxes are validated as ordered integer rectangles inside the image. A question the model cannot localize is marked unresolved and makes the group inapplicable, rather than letting generation proceed on a guessed location.

\textbf{Candidate regions (SP, and MP in-place route).}
From the localized boxes the builder enumerates blank regions of the page as placement candidates. A region is admissible only if it does not overlap any protected evidence box and lies near one: the gap between rectangle edges, divided by the page diagonal, is at most $0.12$. Each region carries a conservative character capacity at a fixed font size. Coverage is recorded per question, and the release does not assume that every question of a group receives a nearby record.

\textbf{Generation.}
Given the localization and the admissible regions, the model writes one candidate for the whole group. On SP it is a single short record per edited page: one printable-ASCII paragraph of at most $320$ characters and at most the chosen region's capacity, carrying only a scoped field and value under a clear heading (e.g.\ \emph{Prior-year record:}), so that it reads as a separate record and never fills an original blank form field or joins an existing table. The scope that excludes it must already be excluded by the question or by the established document identity; an alternate value alone, an invented correction or a ``not the answer'' disclaimer does not qualify. The model also returns, for every question, the existing condition, the added scope and why the addition is excluded. On MP the model takes one of two routes. By default it writes (a) one independent appended page of the same genre, describing a clearly different entity, period or scope with similar fields (title at most $100$ characters, body at most $2{,}400$ characters and $30$ rendered lines), rendered as a separate page placed after the original pages. When any question's referent is the whole document (its title, author, date, or a count of its entries), an appended page would join that referent and create a second valid answer, so the model must instead take (b) the in-place route above, editing one or more existing pages so that page count and document identity are untouched, with every question of the group receiving at least one edited page.

\textbf{Rendering and delivery checks.}
Records are drawn at a fixed font size inside the chosen region; a record that does not fit is rejected, never shrunk. The renderer is then replayed over the untouched original, and the candidate is refused if any pixel outside the record's box differs. For MP appended pages, the original page files and their order are byte-identical, and for each task the builder runs the benchmark's own preparation step and checks that the number of delivered images equals the original count plus one, with the new page last.

\textbf{Review.}
Two independent reviews, each seeing the final rendered images, all questions and their gold answers, audit the group. Each returns group-level checks (natural material, readable, original evidence preserved, safe placement) and per-question checks (evidence supported, answer unchanged, no new correct evidence, no answer hint, scope excluded). Reviewers are instructed to separate a \emph{wrong distractor}, which is the intended treatment, from an \emph{additional correct answer}, which is a defect: a question that asks for 1 January 1982 excludes a clearly labelled 1 January 1981 record, while ``the date of this document'' may become ambiguous once a second date appears. Any false check or uncertainty rejects the candidate. A group may make up to three generation attempts, bounding its logical calls at $1+3\times 3$.

\subsection{Workspace Evidence: SpreadsheetBench, SkillFlow and ContinualSkillBench}

On these sources the gold is not a string that can be shown to a generator: SpreadsheetBench grades against a golden workbook, and SkillFlow and ContinualSkillBench grade the agent's files with native test suites or judges. Generation and review therefore use the public task and the public input files only, and answer-awareness is replaced by an explicit, checkable scope: the generator must return a short \emph{distinction} string that appears verbatim in the material, together with the concrete features it shares with the task.

\textbf{Relation and material.}
Each group is assigned one of three relations by a seeded hash of its identifier: \emph{near-miss} material imitates the topic and structure of the original input and shares concrete features, but visibly differs on a question-relevant constraint (e.g.\ lot codes with prefix \texttt{ZX} when the task filters for \texttt{PK}); \emph{related non-decisive} material shares the domain but supplies no requested evidence or execution advice; \emph{unrelated} material has an independent subject. The model returns a structured note (title, visible scope, columns, rows and optional paragraphs), never raw file bytes, and must not provide the answer, a formula, an answer location, a new instruction or an authoritative correction, nor label the material as irrelevant.

\textbf{Rendering.}
A deterministic renderer writes the note into an existing input file with a format-specific additive operation, selected per group so that every task of the group has an editable input (Table~\ref{tab:nevid_routes}): an \texttt{.xlsx} workbook gains one final worksheet with a name distinct from every original sheet, while the original sheets' XML, cells and formulas remain byte-identical; a \texttt{.csv} file gains one final column with the original rows and cell strings unchanged; a \texttt{.json} file gains one new top-level member with all original keys and values unchanged; \texttt{.docx}, \texttt{.txt} and \texttt{.md} files gain a trailing titled section with the original content retained as a prefix. When no input can be edited safely, or the task has no editable input, a fallback places one separate reference file in the workspace, always related non-decisive, with $2$--$8$ columns, $2$--$12$ rows, $20$--$240$ words and no executable formulas; it does not replace any input or claim to be one.

\textbf{Checks.}
Before any model review, two mechanical gates run. The renderer is replayed over the original file and the result must match the delivered file, which confirms that the original content is intact. A literal gate then rejects any question that counts or enumerates the whole container being appended to, such as the number of worksheets in the workbook, the number of columns in the CSV file, or all top-level keys of the JSON input, since an addition changes that answer even when every original value is preserved. The surviving tasks receive two independent model reviews that see the original and modified excerpts and the actual before/after structure. The first returns four checks (answer preserved, scope distinguishable, no instruction, mechanism fit) and must quote the task's constraints verbatim; it distinguishes input constraints from output constraints, so that a requirement for exactly one output sheet does not forbid an additional input sheet, and it rejects only an actual input-restriction violation, a loss of answer identifiability, a new instruction or hint, or a second equally applicable contradictory fact. The second is an independent ambiguity audit. A rejection by either review makes the group inapplicable; uncertainty or an error leaves it unpublished. The generator may regenerate once after a malformed reply, and a group's logical calls are bounded by $3+2\,|G|$ for a group of $|G|$ tasks.

\subsection{Observation Replay: ALFWorld}

In ALFWorld the space the agent searches is the sequence of observations returned by the simulator, and no text inserted there can be guaranteed consistent with every future state: a sentence about an object or location may become false after any action. N-Evid is therefore constructed without generated content, by replaying genuine public observations. Each group is assigned one of two operators by a seeded hash. \emph{Repeat-visible} appends a copy of lines from the current observation; \emph{history-excerpt} appends lines from the immediately preceding observation. The appended block is labelled \emph{Observation log, step $k$}, with each line quoted, so that it never reads as a current-state assertion. It is delivered after every non-terminal observation, including the initial one when the operator has a source, and is capped at $900$ characters of whole lines. Lines that restate the task goal, instructions or action markup are excluded, so the operator can only repeat environment evidence the agent has already been shown.

The simulator is untouched: the native observation text is preserved as a prefix, and the admissible commands, state transitions, terminal messages and rewards are identical to the clean episode, which is checked mechanically. Because the operator's effect depends on the trajectory, review is run on real deliveries: the builder records a native trace for each task, requires at least one non-reset frame on which the operator actually applied (otherwise the task is marked uncertain and not published), and passes those frames to the same two model reviews used for workspace evidence.

\section{Noise Validity Verification}
\label{app:noise_validation}

Every noisy condition in RSE-Bench is required to satisfy four invariants: the task goal is unchanged, the correct evidence remains reachable within the same interaction budget, the true outcome of every execution is unchanged, and the scoring semantics are unchanged. This appendix describes how each invariant is enforced. Verification runs in three layers of increasing cost and decreasing mechanical decidability. Automated checks come first and make no model calls; model review handles the semantic properties that no program can decide; and a human audit of sampled instances by three independent annotators checks the two materialized releases, N-Inst and N-Evid. A variant enters a release only if it passes the automated and model layers that apply to it; the human audit then re-reviews a random sample of each benchmark's release, and a sampled variant that fails is removed. Table~\ref{tab:validity_layers} gives the mapping.

\begin{table}[t]
\centering
\caption{Where each validity invariant is enforced. \emph{Auto}: deterministic checks without model calls; \emph{Model}: two-sided model review of the final material; \emph{Human}: sampled re-review of each benchmark's release by three independent annotators, a sampled variant passing only if all three pass it. N-Rec and N-Bind are deterministic transformations whose invariants are fully decidable, so they need neither model nor human review.}
\label{tab:validity_layers}
\vspace{6pt}
\small
\setlength{\tabcolsep}{5pt}
\resizebox{\textwidth}{!}{%
\begin{tabular}{@{}lllll@{}}
\toprule
Invariant & N-Inst & N-Evid & N-Rec & N-Bind \\
\midrule
Task goal unchanged           & Auto + Model + Human & Auto + Model + Human & Auto (question never read) & Auto (question never read) \\
Correct evidence reachable    & Model + Human        & Auto (replay) + Model + Human & Auto (execution untouched) & Auto (execution untouched) \\
True outcome unchanged        & Auto (evaluator untouched) & Auto (evaluator untouched) & Auto (score immutable) & Auto (payloads permuted, not edited) \\
Scoring semantics unchanged   & Auto + Model + Human & Auto + Model + Human & Auto & Auto \\
No answer leak or hint        & Auto + Model + Human & Model + Human        & --- & --- \\
\bottomrule
\end{tabular}%
}
\end{table}

\subsection{Automated Verification}

\textbf{Shared release checks.}
Every noisy release is built from, and checked against, the clean base of the same source. After the noise field is removed, each row must be identical to its clean counterpart: same task identifier, question, target, evaluator reference, split and group. Evaluators and scoring assets are copied unchanged, and a release that shadows or rewrites a native scoring asset is refused. Construction and review always write to a new sibling release and never overwrite an existing one, so an audit can compare any published variant with the material it was derived from.

\textbf{N-Inst.}
Each variant is the original question with one insert placed before or after it, the position being fixed by the seed and the task. Five checks run before any model review. (i) \emph{Verbatim question}: the original question must occur verbatim in the noisy prompt, so that nothing of the request is rewritten. (ii) \emph{Non-degenerate insert}: the remaining span must be nonempty and contain at least one full sentence. (iii) \emph{No answer string}: no gold answer string may occur in the insert. For the two DocVQA sources this match respects word boundaries, so that a one-character or numeric answer does not reject every insert containing that character. Sources graded by execution have no answer string and skip this check. (iv) \emph{Source bans}: per-source patterns reject constructions known to change the product, for example a new output language, unit or file name. (v) \emph{No output request}: a deliberately narrow pattern rejects requests to transcribe, copy out, quote verbatim or report the material's content ``along with your answer'', which would move the insert into the graded product. The pattern is narrow on purpose, because unrelated chore requests are the designed content of this noise and a broad list of imperative verbs would reject them silently. For ALFWorld, three further checks are programmatic: any object or receptacle the insert names must appear in the task's own reset observation or goal line (scene membership), statements about listed receptacles may not contradict their presence in the room, and the insert may not request an action on anything the agent could act on within its step budget.

\textbf{N-Evid.}
The construction-specific checks are detailed in Appendix~\ref{app:noise_details}; we summarize the decidable ones here. For image material, the renderer is replayed over the untouched original and every pixel outside the added record's box must be unchanged, protected evidence boxes may not be overlapped, and for appended pages the original page files and their order must be byte-identical, with the task's own preparation step delivering exactly one more image. For file workspaces, replaying the renderer over the original input must reproduce the delivered file, the original sheets, rows, keys or text must be intact, and a literal gate rejects any question that counts or enumerates the whole container that received an addition. For ALFWorld, the native observation must remain a prefix of every delivered observation, the admissible commands, transitions, terminal messages and rewards must equal those of the clean episode, and a recorded native trace must contain at least one non-reset step on which the operator actually applied.

\textbf{N-Rec.}
N-Rec runs on a detached copy of each episode after scoring and never touches the episode that is measured. A configuration is rejected unless it preserves evaluator feedback and citations. At run time the transformation (i) applies only to fit-split episodes that are eligible for learning and appear in the declared task manifest, and passes held-out episodes through unchanged; (ii) edits only string leaves in the declared channels, assistant text and tool output, and a structural comparison of the episode before and after raises an error if any other field, the trajectory structure, an envelope or the multiset of citation tokens changed; (iii) for deletion, verifies that exactly the selected characters were removed and nothing was inserted, with no marker left behind; (iv) for repetition, caps the added characters; and (v) keeps the episode's identity, status, score and learning-eligibility flag immutable, so the condition under which a method decides to update is identical to the clean run. Span selection uses a SHA-256 hash of the seed, source, task and round, so every edit is reproducible and logged with its coordinates.

\textbf{N-Bind.}
N-Bind operates on one fit batch at a time and refuses a batch that crosses source, split, round or skill version, so a misbound outcome always comes from a genuine attempt at the same stage of learning. Before binding, each episode's score must agree with its native metrics, and only dedicated trailing evaluation-feedback rows are treated as movable; an ambiguous row mixing actor content with feedback is refused rather than moved. Binding must be a one-to-one permutation of feedback packets, and a post-condition checks that every execution record and every feedback payload is byte-identical to its original; only their pairing changes. Methods that consume only native evaluator feedback cannot acquire gold through a misbound packet. The resulting mapping records, per slot, the donor attempt and whether the slot is misbound; the per-batch misbinding counts reported in Section~\ref{sec:diagnostics} are read from these records rather than from the configured delay parameters.

\textbf{Exposure check.}
For N-Rec and N-Bind, whether the perturbation actually reached the learner is itself verified. A passive recorder captures every optimizer request, and for a controlled paired replay, with method state, task order and stubbed responses held fixed, the clean and noisy requests are compared call by call. A perturbation counts as exposed only if the prompt changed while the system message and call arguments did not. This is a statement about the learner's inputs, not about model behaviour.

\subsection{Model Review}

Model review decides the properties no program can: whether an insert changes the product, whether an added fact could also answer the question, whether a distractor's scope is visibly excluded. All reviews run at temperature $0$ with thinking disabled, return a fixed JSON schema that is parsed strictly, and are applied to the final material exactly as the agent will see it. A malformed reply, a contradictory verdict (e.g.\ \texttt{ok} with a failed check), an \emph{uncertain} verdict or a transport failure counts as non-acceptance. Reviewers never see any actor's score, since a variant must not be selected by how hard it makes the task. The condensed English renderings below give the decisive criteria; the verbatim prompts, including source-specific blocks, are released with the code.

\begin{tcolorbox}[
    title={Review Prompt: N-Inst (condensed)},
    fonttitle=\bfseries, enhanced, breakable, fontupper=\small,
    colback=gray!10, colframe=black, coltitle=white]
Judge whether the noisy prompt breaks the task relative to the original question.
Background chatter or chore requests unrelated to the task are designed content and are not by themselves grounds for rejection; mentioning a method, tool or concern is not either, unless it is exactly the one needed to solve this task. Reject only if any of the following holds:\\
1.~It introduces a constraint or requirement that would change the answer or the delivered product.\\
2.~It changes the delivery requirements (output language, format, precision, unit, file name, submission method).\\
3.~It asserts a fact that contradicts the reference answer or the task material, or a specific claim the material cannot support.\\
4.~It directly or indirectly leaks the answer: states it, points to where it is located, or names the key method, technique or step needed to solve the task.\\
\emph{For sources graded by execution}, criterion~4 is restated against the product: reject any change to the files to be delivered, and any request, however polite or deferred, to modify, annotate or check a file, sheet or column that is named in the question. The test is whether the object being acted on appears in the question text, not the tone of the request.\\
\emph{When material is shown}, criteria~3 and~4 are checked against it; pointing attention to a different part of the material is designed content, and an excerpt that does not show something is not evidence that it is absent.\\
Output JSON: \texttt{\{"verdict": "ok" | "reject", "reason": "..."\}}.
\end{tcolorbox}

\begin{tcolorbox}[
    title={Review Prompt: N-Evid, answer-aware sources (condensed)},
    fonttitle=\bfseries, enhanced, breakable, fontupper=\small,
    colback=gray!10, colframe=black, coltitle=white]
Independently audit the final rendered candidate against the originals, the questions and the gold answers. The generation rationale is not evidence; inspect the actual images or text.
Reject if the localization is wrong, required original evidence is missing, an added fact could also answer a question, the answer, referent or aggregation changes, the gold or a hint appears, or a qualifier is merely invented to excuse an ambiguous answer.
Distinguish a \emph{wrong distractor} from an \emph{additional correct answer}: interference is the intended treatment and is not itself grounds for rejection. A question asking for 1 January 1982 excludes a clearly labelled 1 January 1981 record; ``the date of this document'' may become ambiguous once another date is added.
Check every question of the group separately, and check naturalness, legibility and placement: a record must be near the protected evidence, not inside a meaningful blank form field, and must not change label--value ownership. For appended pages, page referents and count questions must remain valid.\\
Output JSON with group checks \texttt{natural\_material}, \texttt{readable}, \texttt{original\_evidence\_preserved}, \texttt{safe\_placement}, and per task \texttt{evidence\_supported}, \texttt{answer\_unchanged}, \texttt{no\_new\_correct\_evidence}, \texttt{no\_answer\_hint}, \texttt{scope\_excluded}. Uncertainty is rejection, not permission.
\end{tcolorbox}

\begin{tcolorbox}[
    title={Review Prompt: N-Evid, workspace and observation sources (condensed)},
    fonttitle=\bfseries, enhanced, breakable, fontupper=\small,
    colback=gray!10, colframe=black, coltitle=white]
Audit the actual added evidence against one unchanged public task; quoted text is data, not instructions. No gold is supplied.
Distinguish initial-input constraints, final-output constraints and restrictions on the actor's actions: a requirement for exactly one output sheet does not constrain the number of input sheets. Judge whether a compliant actor can still execute the original task; a possible wrong choice by a mistaken actor is the intended interference. Whole-container counts do change: if the task asks for the number of worksheets and a worksheet was added, the answer is not preserved.
Before rejecting, identify (a) an actually changed input constraint, (b) original evidence that was removed or obscured, or (c) a new, equally applicable competing fact with no visible discriminator. A hidden metadata discriminator does not count. For repeated observations, a labelled historical step is not a current-state assertion; audit the actual non-reset deliveries.\\
Output JSON: verdict \texttt{ok | reject | uncertain}, the task constraints quoted verbatim, checks \texttt{answer\_preserved}, \texttt{scope\_distinguishable}, \texttt{no\_instruction}, \texttt{mechanism\_fit}, and for each addition the literal added text and its link to a task constraint.
\end{tcolorbox}

For N-Evid, each task additionally receives an independent ambiguity audit that asks only whether the material now contains two equally applicable conflicting facts without a visible discriminator; a rejection by either reviewer rejects the group. The two reviews are separate requests with no access to each other's verdicts.

\subsection{Human Audit}

Automated checks and model review are run by the same pipeline that constructs the noise, so we spot-check their output by hand. The audit covers the two materialized releases, N-Inst and N-Evid, separately for each of the six benchmarks. It is a sampled re-review of variants that have already been accepted by the automated and model layers, not an additional gate applied to every variant.

\textbf{Sampling.}
For each benchmark and each of the two noise types, 30 variants are drawn uniformly at random from the accepted releases. For N-Evid the sampling unit is a group, since one piece of material is shared by all questions of the group, and every question of a sampled group is checked.

\textbf{Review.}
Each sampled variant is judged by three annotators from the author team, working independently. Annotators see the clean task, the noisy variant as the agent would receive it (the noisy prompt for N-Inst; the modified page images, input files or delivered observations for N-Evid), and the gold answer where the benchmark has one. They do not see the model reviewers' verdicts or any actor's result on the task. Each annotator answers one question: does the variant keep the original task intact, that is, is the goal and deliverable unchanged, is the correct evidence still present and usable, and does the added content neither supply the answer, a hint or a second correct answer, nor contradict the task's material? A variant passes only if all three annotators answer yes; a single failing judgment counts as a failure, and an annotator who is unsure records a failure.

\clearpage

\section{Case Studies}
\label{app:case_studies}

This appendix expands Section~\ref{sec:case_studies} into eight classes. Each class is stated first as a mechanism, independently of any particular cell of the grid, and is then supported by two concrete cases: one from the ablation setting used throughout Section~\ref{sec:ablation} (SpreadsheetBench, DeepSeek-V4.1-Flash, ACE) and one from a different source, backbone or method, so that no class rests on a single cell.

Figures inside the boxes are absolute task counts from the audit of one complete run of the pipeline, under that run's per-source scoring key, and not the run-averaged percentages of Table~\ref{tab:main_results1}. Counts are therefore comparable within a case but not across the two cases of a class: the cross-setting case changes the source, so the scoring key and the number of tasks change with it (SpreadsheetBench scores a hard restriction over 80 tasks, SkillFlow a reward threshold over 50). Single-task differences remain inside run-to-run variation (Table~\ref{tab:run_variability}); every class is claimed at the level of a set of tasks, and the cases are shown to expose the mechanism rather than to establish significance on their own. Two further observations that constrain how the per-cell tables may be read --- that a method can be untouched by a noise position for lack of an input channel that carries it, and that the same position reverses sign across source, backbone and method --- are reported with the diagnostics in Section~\ref{sec:ablation} rather than as case classes.

\subsection{Learned Skills and Noise Are Mutually Exclusive}
\label{app:case_interaction}

A self-evolved skill is a compression of the regularities of the environment it was learned in, and it inevitably freezes in a set of premises about that environment's surface structure. Noise changes exactly that surface structure while the task goal itself is left intact, so the premises inside the skill stop holding and the agent goes on acting on one that is no longer true. This failure differs in form from the task becoming harder: on the same item, a model carrying no skill store often still solves it from general competence, and the loss appears only on the side that enters the noisy environment carrying what it learned. Such losses should therefore be attributed to a conflict between the learned prior and the present environment rather than to a rise in difficulty; where the noise is strong enough, a skill store can be a net liability.

\begin{tcolorbox}[
    title={Case Study: Learned Skills and Noise Are Mutually Exclusive},
    fonttitle=\bfseries,
    enhanced,
    breakable,
    fontupper=\small,
    colback=gray!10,
    colframe=black,
    coltitle=white,
    colbacktitle=black
]
\textbf{Case 1} --- SpreadsheetBench, DeepSeek-V4.1-Flash, ACE; $N_{\mathrm{Evid}}$ on the test release; task \texttt{33157}.

The unevolved agent scores $1$ on this noisy item. The cleanly evolved playbook scores $1$ on the clean release of the same task and $0$ on the noisy one. The noisy run ends in state \texttt{completed}, so nothing was truncated, and the trajectory is shorter rather than longer ($21 \rightarrow 13$ steps): the agent finished early on a premise the distractor file had already invalidated. Over the whole arm the unevolved baseline is essentially flat when the noise is moved into the test release, while the cleanly evolved store moves $11$ down and $1$ up under $N_{\mathrm{Inst}}$, and $10$ down and $1$ up under $N_{\mathrm{Evid}}$.

\vspace{0.8em}
\tcbline
\vspace{0.8em}

\textbf{Case 2} --- SkillFlow, GPT-5.2, ACE; $N_{\mathrm{Evid}}$ on the test release; task \texttt{Document-Fraud-Detection/clinical-trial-participant-release-audit}.

The same shape recurs on another source and another backbone, in a more extreme form. On this task: unevolved $1$, cleanly evolved on the clean release $1$, cleanly evolved on the noisy release $0$. Over the arm, $6$ tasks drop and $1$ rises, and all $6$ of the dropped tasks are solved by the unevolved agent on the same noisy items, so they were not unsolvable. Aggregated over the release, the unevolved agent ($16/50$) outscores the cleanly evolved one ($9/50$), against $14/50$ for the evolved store on the clean release. Under this amount of evidence noise the skill store is a net liability rather than a diminished asset.
\end{tcolorbox}

\subsection{Precautionary Clauses Consume the Interaction Budget}
\label{app:case_budget}

Noise leads a method to learn cautious discipline of the form ``enumerate first, verify first, only then act''. Such discipline costs almost nothing in a clean environment, which is why it always looks safe at learning time; but once it is written down as an unconditional preamble, it charges a fixed toll on every task. When the environment contains more objects to be checked --- the extra files, sheets and paragraphs all have to be read first --- that toll scales with the number of objects, and a bounded interaction budget is spent before the real edit is ever attempted. The failure therefore presents as a step or output limit rather than as a wrong answer: its cause lies in the skill layer, but it is recorded as a resource problem in the execution layer.

\begin{tcolorbox}[
    title={Case Study: Precautionary Clauses Consume the Interaction Budget},
    fonttitle=\bfseries,
    enhanced,
    breakable,
    fontupper=\small,
    colback=gray!10,
    colframe=black,
    coltitle=white,
    colbacktitle=black
]
\textbf{Case 1} --- SpreadsheetBench, DeepSeek-V4.1-Flash, ACE; $N_{\mathrm{Inst}}$ on the test release; task \texttt{14240}.

The clean run of this task terminates normally in $13$ steps; with the inserted sentences present the same store runs $40$ steps and ends at \texttt{step\_limit}, never submitting. The unevolved agent scores $1$ on the same noisy item. The playbook in force front-loads clauses such as enumerating every worksheet and reconciling the declared structure against the actual one before writing anything, and the inserted text multiplies the surfaces that clause demands be checked. Arm-wide, $2$ of the $11$ tasks that drop under $N_{\mathrm{Inst}}$ and $4$ of the $10$ that drop under $N_{\mathrm{Evid}}$ end in a step or output limit, against an arm mean of only $8.5$--$13.2$ steps.

\vspace{0.8em}
\tcbline
\vspace{0.8em}

\textbf{Case 2} --- SkillFlow, DeepSeek-V4.1-Flash, SkillOpt; $N_{\mathrm{Evid}}$; task \texttt{Compensation-Scenario-Modeling/08\_construction\_union\_model}.

The same task takes $22$ steps in the clean arm, which solves it, and $40$ steps here, ending at \texttt{step\_limit} with no submission. This setting also supplies the third reading the ablation setting cannot: $9/50$ tasks exhaust the budget in this arm against $3/50$ in the cleanly evolved arm, and when the \emph{same} frozen store (v3, $6354$ characters, $4$ front-loaded clauses) is re-evaluated on the clean release it still exhausts $6/50$. The proximate cause is thus the precautionary clause in the store, with the noise only triggering it more often. The class remains a minority phenomenon --- $6$ cases in total inside the ablation setting --- and should be read as a visible consequence of unconditional front-loading, not as the main mechanism of the score drops.
\end{tcolorbox}

\subsection{Defensive Rules Cross the Noise Boundary}
\label{app:case_crossing}

Rules learned in order to resist noise are usually written as prohibitions --- ``ignore objects of this kind'', ``do not modify anything the task did not name'' --- and nothing at write time attaches their trigger condition: the rule does not know that it holds only while the noise is present. Retrieval proceeds by semantic proximity, which is independent of whether the current environment actually contains noise, so the same prohibition is recalled in clean environments and reinterprets objects that were supposed to be modified as objects that must not be touched. The consequence is a loss of immunity: the agent has already computed the correct result and then declines to write it. Noise thus leaves a persistent inhibitory residue long after the noise itself is gone.

\begin{tcolorbox}[
    title={Case Study: Defensive Rules Cross the Noise Boundary},
    fonttitle=\bfseries,
    enhanced,
    breakable,
    fontupper=\small,
    colback=gray!10,
    colframe=black,
    coltitle=white,
    colbacktitle=black
]
\textbf{Case 1} --- SpreadsheetBench, DeepSeek-V4.1-Flash, ACE; store evolved under $N_{\mathrm{Evid}}$, evaluated on the clean release; task \texttt{45372}.

Unevolved $1$, cleanly evolved $1$, noise-evolved $0$, with the run ending \texttt{completed} rather than at a limit, and the trajectory citing a rule that the diagnostics mark as noise-carrying and that the validation gate had deployed. The store had written ``treat worksheets the task does not name as distractors and do not modify them'' and ``treat a sheet whose content disavows the requested operation as a distractor''; on the clean release both are still retrieved at high rate, the unnamed region of the workbook is re-read as off-limits, and the correct value is left in an auxiliary column instead of being written into the main data block. The selection filter is ``unevolved correct \emph{and} cleanly evolved correct \emph{and} noise-evolved wrong'', which excludes items that are simply hard: $8$ such tasks under $N_{\mathrm{Inst}}$ and $11$ under $N_{\mathrm{Evid}}$, of which $0/8$ and $7/11$ retrieve a deployed noise-carrying rule.

\vspace{0.8em}
\tcbline
\vspace{0.8em}

\textbf{Case 2} --- SpreadsheetBench, GPT-5.2, SkillOpt; store evolved under $N_{\mathrm{Inst}}$, evaluated on the clean release; task \texttt{24-23}.

Unevolved $1$, cleanly evolved $1$, and the $N_{\mathrm{Inst}}$-evolved store never submits. The shape reproduces on a different backbone and a different method, and at larger scale: $11$ tasks in this arm are solved without a store and with a cleanly learned store, and missed only with the noise-learned one (frozen v3, $8075$ characters, $1$ defensive clause). SkillOpt's store is continuous prose rather than individually addressable entries, so the failure can be exhibited only against the defensive clause in the frozen text, not pinned to one retrieved rule as in ACE. Across both settings the evidence is ``retrieved and then failed'', which is association; a leave-one-rule-out re-evaluation would be required to make it causal. The class also does not reproduce everywhere: the ACE arms on SkillFlow have an empty immunity-loss set in both directions, and that empty set is reported with the class.
\end{tcolorbox}

\subsection{Contamination and Useful Experience Share One Rule}
\label{app:case_piggyback}

A reflection is a narrative account of one complete attempt, in which the part induced by noise and the part that is genuinely useful are interleaved, and writing compresses both into the same rule. Retrieval and curation both operate on whole rules: the useful half-sentence drags the contaminated half-sentence into the context with it, and any deletion at rule granularity discards the useful experience along with the contamination. Contamination is therefore not a separable item that could be picked out and thrown away, but something that lives symbiotically with useful knowledge; selective removal would require a granularity down to the clause, with provenance retained per clause.

\begin{tcolorbox}[
    title={Case Study: Contamination and Useful Experience Share One Rule},
    fonttitle=\bfseries,
    enhanced,
    breakable,
    fontupper=\small,
    colback=gray!10,
    colframe=black,
    coltitle=white,
    colbacktitle=black
]
\textbf{Case 1} --- SpreadsheetBench, DeepSeek-V4.1-Flash, ACE; $N_{\mathrm{Evid}}$; rule \texttt{general-00001}.

Written in round $1$, passed the validation gate into the frozen version, cited on $75$ of the $80$ test items, of which $56$ are still answered correctly --- the rule is useful and contaminated at once, and the diagnostics flag it as a piggyback: one clause is a precautionary discipline the clean arm learns as well, the other carries the distractor. Noise-carrying rules per position in this arm are $3/30$ under $N_{\mathrm{Inst}}$, $3/22$ under $N_{\mathrm{Evid}}$, $4/23$ under $N_{\mathrm{Rec}}$ and $12/28$ under $N_{\mathrm{Bind}}$, and the piggyback subset is retrieved at a visibly higher rate than purely noise-derived rules, because the high rate is bought by the useful clause. With curation at rule granularity --- the granularity ACE's interface exposes --- no operation deletes only the contaminated clause.

\vspace{0.8em}
\tcbline
\vspace{0.8em}

\textbf{Case 2} --- SkillFlow, DeepSeek-V4.1-Flash, ACE; $N_{\mathrm{Inst}}$; rule \texttt{general-00005}.

Written in round $01$ and carried into frozen v1 ($10$ rules), this rule tells the agent to separate narrative distraction from the machine-readable task: its first clause is elicited by the inserted sentence, its second clause is a genuine requirement of the SkillFlow task. It is retrieved on $44$ of $50$ test items, $21$ of which are still answered correctly. Two caveats travel with the class. Noise-carrying and piggyback labels come from the keyword detectors in the diagnostics and carry false positives, for which the clean arm's hit rate is the floor; and this particular piggyback is identified by reading the text --- the automatic front-loading-and-guard detector does not fire on it, so the detector covers only one of the spellings in which the two clauses can be fused.
\end{tcolorbox}

\subsection{Misbinding Induces Fabrication That Is Then Frozen In}
\label{app:case_fabrication}

When the pairing between an outcome and the process that produced it is broken, the method reads evidence that contradicts itself: the process shows the work being done, the verdict says it failed. The interface gives it no way to question the pairing itself, so it instead constructs an explanation for a failure that never happened and writes that explanation down as a rule. The error crosses from the feedback layer into the knowledge layer. If that round happens to pass validation, the invented rule is deployed and retrieved repeatedly thereafter, and a one-off perturbation of the binding has been converted into a persistent false prior.

\begin{tcolorbox}[
    title={Case Study: Misbinding Induces Fabrication That Is Then Frozen In},
    fonttitle=\bfseries,
    enhanced,
    breakable,
    fontupper=\small,
    colback=gray!10,
    colframe=black,
    coltitle=white,
    colbacktitle=black
]
\textbf{Case 1} --- SpreadsheetBench, DeepSeek-V4.1-Flash, ACE; $N_{\mathrm{Bind}}$, round $01$; task \texttt{41978}.

The task truly scores $1$; what is presented to the method is the donor attempt's non-$1$ outcome (donor \texttt{227-40}, true score $0$). The reflection duly reads the episode as a failure, supplies a mechanism for it, and that mechanism becomes the stated reason for $2$ skill writes. Over the arm's three rounds, $32$ payloads are swapped, $16$ of them presenting a true $1$ as non-$1$, and the reflections on those items consistently narrate a success as a failure.

\vspace{0.8em}
\tcbline
\vspace{0.8em}

\textbf{Case 2} --- SkillFlow, DeepSeek-V4.1-Flash, ACE; $N_{\mathrm{Bind}}$, round $01$; task \texttt{Distribution-Center-Auditing/harbor\_trailer\_detention\_audit}.

This setting supplies the missing half. The task truly scores $1$ and is presented to the method as $0.0$; $6$ of the $50$ reflections in this round are fed swapped scores ($36$ of $150$ queue entries had their payload replaced across the run). The round is then accepted by the gate (\texttt{accept\_new\_best}, promoted) and becomes frozen v1, whose $11$ rules all originate in it; the most-retrieved of them is cited on $50$ of $50$ test items. The full chain --- misbinding, fabrication, write, deployment, retrieval --- is therefore observable, but in a second form: the stated reason for the write is an aggregate claim over the round (``nearly every $0.0$ reflection says \ldots''), and the fabricated failure enters its denominator, so it cannot be traced back to any single rule even though it is frozen into the store. The two forms should be reported separately.
\end{tcolorbox}

\subsection{Attribution Is Displaced from the Error to the Observation Channel}
\label{app:case_displacement}

When the record is mutilated, the method loses the basis on which a failure could be localised, yet it is still required to produce a cause. The missing evidence does not lead it to delete or shorten steps; it displaces the attribution onto the only visible anomaly --- that the observation itself cannot be trusted. The rules it then writes are meta-rules about the credibility of logs and tools, unrelated to task competence and unable to improve behaviour on later tasks. Such displacement still looks like learning from the outside, but what is learned is suspicion of the observation channel, and it goes on occupying room in the store that task knowledge should hold.

\begin{tcolorbox}[
    title={Case Study: Attribution Is Displaced from the Error to the Observation Channel},
    fonttitle=\bfseries,
    enhanced,
    breakable,
    fontupper=\small,
    colback=gray!10,
    colframe=black,
    coltitle=white,
    colbacktitle=black
]
\textbf{Case 1} --- SpreadsheetBench, DeepSeek-V4.1-Flash, ACE; $N_{\mathrm{Rec}}$, round $01$; task \texttt{22-47}.

The task truly scores $0$. The reflection attributes the failure to garbled or truncated output, does not claim that any step was unnecessary, and produces $1$ skill write about the trustworthiness of tool output. Arm-wide, $94/150 = 63\%$ of reflections hit the garbled-or-truncated detector, against a false-positive floor of $11/150 = 7\%$ measured with the same detector on the clean arm. The counter-evidence holds simultaneously: ``some step was unnecessary'' fires on $3/150$ here and $9/150$ in the clean arm, so it is no higher than the floor.

\vspace{0.8em}
\tcbline
\vspace{0.8em}

\textbf{Case 2} --- SkillFlow, DeepSeek-V4.1-Flash, ACE; $N_{\mathrm{Rec}}$, round $01$; task \texttt{Cross-Format-Data-Reconciliation/01-cloud-service-portfolio-diff}.

The displacement is stronger on this source. The task truly scores $1$, and the reflection reports the trace as garbled. Arm-wide the detector fires on $119/150 = 79\%$ against a clean-arm floor of $20/150 = 13\%$, a gap of $66$ points, and $2$ entries written in this round put log untrustworthiness into their content or their stated reason. Over the same reflections, ``some step was unnecessary'' moves only from $28/150 = 19\%$ to $34/150 = 23\%$. The two gaps differ by an order of magnitude, so the dominant change is a displacement of attribution onto the record and not a proportionate tendency to skip steps. The corrupted view is not stored separately --- $N_{\mathrm{Rec}}$ is replayed from a seed at read time --- so the material for both cases is the undamaged record of the same episode alongside the reflection text.
\end{tcolorbox}

\subsection{The Validation Gate Is the Only Filter}
\label{app:case_gate}

In a pipeline of this shape, curation accepts essentially everything that is written, and the only real selection happens when a version is chosen against the validation set. Noise therefore has exactly two outlets: where the round passes, contamination is deployed together with the useful rules; where it does not, the whole round is rolled back, possibly all the way to an empty store. The two outcomes are indistinguishable in the test score --- a score that did not drop may mean the contamination was resisted, or that nothing was learned at all. Separating them requires reporting ``whether learning happened'' and ``what the test score was'' as two distinct quantities.

\begin{tcolorbox}[
    title={Case Study: The Validation Gate Is the Only Filter},
    fonttitle=\bfseries,
    enhanced,
    breakable,
    fontupper=\small,
    colback=gray!10,
    colframe=black,
    coltitle=white,
    colbacktitle=black
]
\textbf{Case 1} --- SpreadsheetBench, DeepSeek-V4.1-Flash, ACE; the four noise arms and their gate records.

ACE's curator is \texttt{AlwaysAccept}, so every proposed rule is written and the version selection is the only interception point. Both outcomes are visible in one arm set. Contamination ships: the $N_{\mathrm{Evid}}$ arm writes $22$ rules, freezes at v3 with all $22$ present, $3$ of them flagged as noise-carrying and all $3$ deployed; the $N_{\mathrm{Inst}}$ arm writes $30$, freezes at v2 and keeps $18$. The round is discarded: the $N_{\mathrm{Bind}}$ arm freezes at v0, so what is deployed is an empty playbook.

\vspace{0.8em}
\tcbline
\vspace{0.8em}

\textbf{Case 2} --- SkillFlow, GPT-5.2, ACE; the same four arms.

Here the dichotomy is sharper, and it is invisible in the scores. Three of the four noise arms roll back to v0. Read from the score column alone, three cells in which no learning survived look at least as healthy as the clean cell. Freezing at v0 is a fact about whether learning happened and not a robustness result, which makes reporting the frozen version alongside any per-position robustness table a requirement rather than a suggestion.
\end{tcolorbox}

\subsection{Store Size Drifts Toward Both Extremes}
\label{app:case_drift}

Since filtering happens at that one point, the size of the skill store drifts along it toward both extremes: where a round is let through, the store inflates along with its defensive clauses and transfers the cost to the execution layer; where it is not, the whole round is rolled back and the store ends smaller than it would have been without noise, or empty. Whether a skill store grows or shrinks under noise is thus not a property of the noise position but a consequence of whether that round passed validation. Reading size in one direction only, as inflation, hides the more serious failure at the other end: an entire round of learning is discarded, and it leaves almost no trace in the score.

\begin{tcolorbox}[
    title={Case Study: Store Size Drifts Toward Both Extremes},
    fonttitle=\bfseries,
    enhanced,
    breakable,
    fontupper=\small,
    colback=gray!10,
    colframe=black,
    coltitle=white,
    colbacktitle=black
]
\textbf{Case 1} --- SpreadsheetBench, DeepSeek-V4.1-Flash, ACE; frozen store sizes by arm.

The clean arm freezes on an earlier version ($11$ rules, $2730$ characters, $0$ clauses with defensive wording). The noise arms advance further and inflate, and what they add is dominated by clauses of the form ``treat X as a distractor'' and ``reconcile everything before acting'': $N_{\mathrm{Inst}}$ freezes at $18$ rules, $6330$ characters, $2$ defensive clauses, and $N_{\mathrm{Evid}}$ at v3 with $22$ rules, $8279$ characters, $2$ defensive clauses. This is the path that connects directly to budget exhaustion in the execution layer (Section~\ref{app:case_budget}).

\vspace{0.8em}
\tcbline
\vspace{0.8em}

\textbf{Case 2} --- SkillFlow, DeepSeek-V4.1-Flash, ACE; the same method and backbone on another source.

The direction reverses. The clean arm freezes at v3 with $23$ rules and $9752$ characters, and all four noise arms write comparably but end \emph{smaller}: $N_{\mathrm{Inst}}$ at v1 with $10$ rules / $4144$ characters (from $25$ written), $N_{\mathrm{Evid}}$ at v0 with $0$ rules / $21$ characters (from $23$ written), $N_{\mathrm{Rec}}$ at v2 with $19$ rules / $5406$ characters (from $29$ written), and $N_{\mathrm{Bind}}$ at v1 with $11$ rules / $4293$ characters (from $24$ written). Taken together, the two cases license only the weaker statement: noise does not reliably enlarge a skill store, it pushes the size to one extreme or the other, and which extreme is decided by the validation gate (Section~\ref{app:case_gate}) rather than by the noise position. Rule and character counts are descriptive statistics and carry no causal claim, and the defensive-wording label comes from a keyword detector with false positives.
\end{tcolorbox}

\begin{table}[t]
  \centering
  \captionsetup{position=top, skip=8pt}
  \caption{Noise process diagnostics for ACE and RethinkSkill on SpreadsheetBench (DeepSeek-V4.1-Flash). \emph{IER} is the proportion of fit-task optimizer inputs containing an injected perturbation. \emph{Selected version} ($v^{*}$) is the skill version selected on the validation set for final testing, with $v0$ denoting the initial state. \emph{CCS} is the mean term-frequency cosine similarity between candidates from a noise condition and candidates from a clean run. The bracketed reference, $\mathrm{CCS}_{\mathrm{ref}}$, is the mean CCS over all $\binom{K}{2}$ pairs of $K$ independent clean runs for each method, computed using the same cross-run procedure. A dash indicates an undefined or inapplicable quantity. CCS values are comparable within a method.}
  \label{tab:noise_diagnostics}
  \small
  \setlength{\tabcolsep}{5pt}
  \begin{tabular}{llccc}
  \toprule
  Method & Noise & IER & Selected version ($v^{*}$) & CCS [ref] \\
  \midrule
  \multirow{4}{*}{ACE}
  & $N_{\mathrm{Inst}}$ & $100\%$ $(150/150)$ & $v2$ & $0.764\,[0.814]$ \\
  & $N_{\mathrm{Evid}}$ & $97\%$ $(146/150)$  & $v3$ & $0.702\,[0.814]$ \\
  & $N_{\mathrm{Rec}}$  & $100\%$ $(150/150)$ & $v1$ & $0.775\,[0.814]$ \\
  & $N_{\mathrm{Bind}}$ & $21\%$ $(32/150)$   & $v2$ & $0.727\,[0.814]$ \\
  \midrule
  \multirow{4}{*}{RethinkSkill}
  & $N_{\mathrm{Inst}}$ & --                  & $v1$ & $0.820\,[0.901]$ \\
  & $N_{\mathrm{Evid}}$ & --                  & $v2$ & $0.826\,[0.901]$ \\
  & $N_{\mathrm{Rec}}$  & $100\%$ $(150/150)$ & $v3$ & $0.844\,[0.901]$ \\
  & $N_{\mathrm{Bind}}$ & $15\%$ $(22/150)$   & $v1$ & $0.783\,[0.901]$ \\
  \bottomrule
  \end{tabular}
\end{table}

\begin{table*}[p]
\centering
\caption{Full results on RSE-Bench (\%, means over three independent runs). Subscripts show relative changes, $100(s-b)/b$, from the corresponding $C\!\to\!C$ baseline $b$: green $\downarrow$ for decreases, gray $\uparrow$ for increases, and $\leftrightarrow$ for no change. Columns specify learning $\rightarrow$ evaluation; $C$ is clean, and Inst, Evid, Rec, and Bind denote instruction, evidence, record, and binding noise ($N_{\mathrm{Inst}}$--$N_{\mathrm{Bind}}$). Each benchmark block reports that benchmark's split directly; the RSE-Bench (Avg) block averages scores equally over all six benchmarks before computing rates. Underlines mark the largest relative decrease in each noise-condition column across the six method--model combinations of its own block, using unrounded rates and including exact ties, with the same rule applied in the Avg block. Per-run standard deviations appear in Table~\ref{tab:run_variability}.}
\label{tab:main_results}
\vspace{10pt}
\begingroup
\fontsize{9}{10.4}\selectfont
\setlength{\tabcolsep}{1.6pt}
\renewcommand{\arraystretch}{1.12}
\newcommand{\rsebase}[1]{$#1$}
\newcommand{\rserate}[2]{%
  \begingroup
  \def\rsearrow{#1}%
  \def\rsedown{\downarrow}%
  \def\rseup{\uparrow}%
  \ifx\rsearrow\rsedown
    \color{green!50!black}%
  \else
    \ifx\rsearrow\rseup
      \color{black!55}%
    \else
      \color{black}%
    \fi
  \fi
  #1#2\%%
  \endgroup
}
\newcommand{\rsecell}[3]{%
  $#1_{\scalebox{0.78}{$\rserate{#2}{#3}$}}$%
}
\newcommand{\rseworst}[3]{%
  $\underline{#1}_{\scalebox{0.78}{$\rserate{#2}{#3}$}}$%
}
\resizebox{\textwidth}{!}{%
\begin{tabular}{@{}ll*{8}{c}@{}}
\toprule
\multirow{2}{*}{Method} & \multirow{2}{*}{Model}
& \multicolumn{2}{c}{Reference}
& \multicolumn{4}{c}{Noisy learning}
& \multicolumn{2}{c}{Noisy evaluation} \\
\cmidrule(lr){3-4}\cmidrule(lr){5-8}\cmidrule(l){9-10}
& & Initial & $C\!\to\!C$ & $N_{\mathrm{Inst}}\!\to\!C$ & $N_{\mathrm{Evid}}\!\to\!C$
& $N_{\mathrm{Rec}}\!\to\!C$ & $N_{\mathrm{Bind}}\!\to\!C$ & $C\!\to\!N_{\mathrm{Inst}}$ & $C\!\to\!N_{\mathrm{Evid}}$ \\
\midrule
\multicolumn{10}{@{}l}{\textbf{SpreadsheetBench}} \\
\multirow{2}{*}{ACE} & DeepSeek-V4.1-Flash & \rsebase{79.58} & \rsebase{90.00} & \rsecell{74.58}{\downarrow}{17.1} & \rseworst{74.17}{\downarrow}{17.6} & \rsecell{83.33}{\downarrow}{7.4} & \rsecell{80.00}{\downarrow}{11.1} & \rseworst{77.50}{\downarrow}{13.9} & \rseworst{78.75}{\downarrow}{12.5} \\
 & GPT-5.2 & \rsebase{45.00} & \rsebase{40.83} & \rsecell{44.58}{\uparrow}{9.2} & \rsecell{40.00}{\downarrow}{2.0} & \rsecell{42.50}{\uparrow}{4.1} & \rsecell{46.67}{\uparrow}{14.3} & \rsecell{42.08}{\uparrow}{3.1} & \rsecell{37.08}{\downarrow}{9.2} \\
\multirow{2}{*}{RethinkSkill} & DeepSeek-V4.1-Flash & \rsebase{79.58} & \rsebase{88.33} & \rsecell{80.00}{\downarrow}{9.4} & \rsecell{78.75}{\downarrow}{10.8} & \rsecell{82.92}{\downarrow}{6.1} & \rsecell{89.17}{\uparrow}{1.0} & \rsecell{81.67}{\downarrow}{7.5} & \rsecell{81.67}{\downarrow}{7.5} \\
 & GPT-5.2 & \rsebase{45.00} & \rsebase{45.83} & \rsecell{39.58}{\downarrow}{13.6} & \rsecell{40.00}{\downarrow}{12.7} & \rsecell{52.08}{\uparrow}{13.6} & \rseworst{39.17}{\downarrow}{14.5} & \rsecell{44.58}{\downarrow}{2.7} & \rsecell{45.00}{\downarrow}{1.8} \\
\multirow{2}{*}{SkillOpt} & DeepSeek-V4.1-Flash & \rsebase{79.58} & \rsebase{87.08} & \rsecell{86.25}{\downarrow}{1.0} & \rsecell{81.25}{\downarrow}{6.7} & \rsecell{83.75}{\downarrow}{3.8} & \rsecell{97.08}{\uparrow}{11.5} & \rsecell{86.25}{\downarrow}{1.0} & \rsecell{87.50}{\uparrow}{0.5} \\
 & GPT-5.2 & \rsebase{45.00} & \rsebase{47.92} & \rseworst{33.75}{\downarrow}{29.6} & \rsecell{46.25}{\downarrow}{3.5} & \rseworst{39.58}{\downarrow}{17.4} & \rsecell{45.42}{\downarrow}{5.2} & \rsecell{50.42}{\uparrow}{5.2} & \rsecell{47.08}{\downarrow}{1.8} \\
\midrule
\multicolumn{10}{@{}l}{\textbf{SP-DocVQA}} \\
\multirow{2}{*}{ACE} & DeepSeek-V4.1-Flash & \rsebase{92.92} & \rsebase{92.50} & \rsecell{92.50}{\leftrightarrow}{0.0} & \rsecell{89.58}{\downarrow}{3.2} & \rsecell{93.33}{\uparrow}{0.9} & \rsecell{93.75}{\uparrow}{1.4} & \rsecell{90.42}{\downarrow}{2.2} & \rsecell{92.92}{\uparrow}{0.5} \\
 & GPT-5.2 & \rsebase{88.33} & \rsebase{90.83} & \rsecell{91.25}{\uparrow}{0.5} & \rsecell{89.58}{\downarrow}{1.4} & \rseworst{85.00}{\downarrow}{6.4} & \rsecell{89.17}{\downarrow}{1.8} & \rsecell{86.25}{\downarrow}{5.0} & \rsecell{88.33}{\downarrow}{2.8} \\
\multirow{2}{*}{RethinkSkill} & DeepSeek-V4.1-Flash & \rsebase{92.92} & \rsebase{97.92} & \rsecell{97.50}{\downarrow}{0.4} & \rsecell{96.25}{\downarrow}{1.7} & \rsecell{96.25}{\downarrow}{1.7} & \rseworst{93.75}{\downarrow}{4.3} & \rsecell{94.17}{\downarrow}{3.8} & \rseworst{91.67}{\downarrow}{6.4} \\
 & GPT-5.2 & \rsebase{88.33} & \rsebase{90.83} & \rsecell{92.08}{\uparrow}{1.4} & \rsecell{88.75}{\downarrow}{2.3} & \rsecell{89.58}{\downarrow}{1.4} & \rsecell{91.67}{\uparrow}{0.9} & \rsecell{87.92}{\downarrow}{3.2} & \rsecell{92.92}{\uparrow}{2.3} \\
\multirow{2}{*}{SkillOpt} & DeepSeek-V4.1-Flash & \rsebase{92.92} & \rsebase{96.25} & \rsecell{96.67}{\uparrow}{0.4} & \rsecell{96.67}{\uparrow}{0.4} & \rsecell{96.67}{\uparrow}{0.4} & \rsecell{95.42}{\downarrow}{0.9} & \rsecell{95.42}{\downarrow}{0.9} & \rsecell{95.42}{\downarrow}{0.9} \\
 & GPT-5.2 & \rsebase{88.33} & \rsebase{94.17} & \rseworst{92.08}{\downarrow}{2.2} & \rseworst{89.58}{\downarrow}{4.9} & \rsecell{88.33}{\downarrow}{6.2} & \rsecell{93.33}{\downarrow}{0.9} & \rseworst{89.17}{\downarrow}{5.3} & \rsecell{90.42}{\downarrow}{4.0} \\
\midrule
\multicolumn{10}{@{}l}{\textbf{MP-DocVQA}} \\
\multirow{2}{*}{ACE} & DeepSeek-V4.1-Flash & \rsebase{67.08} & \rsebase{71.25} & \rsecell{68.75}{\downarrow}{3.5} & \rseworst{63.75}{\downarrow}{10.5} & \rsecell{68.75}{\downarrow}{3.5} & \rsecell{65.83}{\downarrow}{7.6} & \rsecell{66.67}{\downarrow}{6.4} & \rsecell{67.08}{\downarrow}{5.9} \\
 & GPT-5.2 & \rsebase{68.33} & \rsebase{70.00} & \rsecell{69.17}{\downarrow}{1.2} & \rsecell{71.67}{\uparrow}{2.4} & \rsecell{68.33}{\downarrow}{2.4} & \rsecell{67.50}{\downarrow}{3.6} & \rsecell{68.75}{\downarrow}{1.8} & \rsecell{65.83}{\downarrow}{6.0} \\
\multirow{2}{*}{RethinkSkill} & DeepSeek-V4.1-Flash & \rsebase{67.08} & \rsebase{69.58} & \rsecell{67.92}{\downarrow}{2.4} & \rsecell{65.00}{\downarrow}{6.6} & \rseworst{65.00}{\downarrow}{6.6} & \rseworst{63.75}{\downarrow}{8.4} & \rsecell{69.58}{\leftrightarrow}{0.0} & \rsecell{62.92}{\downarrow}{9.6} \\
 & GPT-5.2 & \rsebase{68.33} & \rsebase{75.83} & \rseworst{70.42}{\downarrow}{7.1} & \rsecell{67.92}{\downarrow}{10.4} & \rsecell{71.25}{\downarrow}{6.0} & \rsecell{70.42}{\downarrow}{7.1} & \rseworst{69.17}{\downarrow}{8.8} & \rseworst{64.17}{\downarrow}{15.4} \\
\multirow{2}{*}{SkillOpt} & DeepSeek-V4.1-Flash & \rsebase{67.08} & \rsebase{69.17} & \rsecell{67.08}{\downarrow}{3.0} & \rsecell{70.00}{\uparrow}{1.2} & \rsecell{68.33}{\downarrow}{1.2} & \rsecell{71.25}{\uparrow}{3.0} & \rsecell{69.58}{\uparrow}{0.6} & \rsecell{59.17}{\downarrow}{14.5} \\
 & GPT-5.2 & \rsebase{68.33} & \rsebase{67.50} & \rsecell{69.58}{\uparrow}{3.1} & \rsecell{68.33}{\uparrow}{1.2} & \rsecell{68.75}{\uparrow}{1.9} & \rsecell{69.58}{\uparrow}{3.1} & \rsecell{69.58}{\uparrow}{3.1} & \rsecell{63.33}{\downarrow}{6.2} \\
\bottomrule
\end{tabular}%
}
\endgroup
\end{table*}

\begin{table*}[p]
\ContinuedFloat
\centering
\caption[]{Full results on RSE-Bench (continued). The notation and reporting conventions are the same as in the first part of Table~\ref{tab:main_results}.}
\vspace{10pt}
\begingroup
\fontsize{9}{10.4}\selectfont
\setlength{\tabcolsep}{1.6pt}
\renewcommand{\arraystretch}{1.12}
\newcommand{\rsebase}[1]{$#1$}
\newcommand{\rserate}[2]{%
  \begingroup
  \def\rsearrow{#1}%
  \def\rsedown{\downarrow}%
  \def\rseup{\uparrow}%
  \ifx\rsearrow\rsedown
    \color{green!50!black}%
  \else
    \ifx\rsearrow\rseup
      \color{black!55}%
    \else
      \color{black}%
    \fi
  \fi
  #1#2\%%
  \endgroup
}
\newcommand{\rsecell}[3]{%
  $#1_{\scalebox{0.78}{$\rserate{#2}{#3}$}}$%
}
\newcommand{\rseworst}[3]{%
  $\underline{#1}_{\scalebox{0.78}{$\rserate{#2}{#3}$}}$%
}
\resizebox{\textwidth}{!}{%
\begin{tabular}{@{}ll*{8}{c}@{}}
\toprule
\multirow{2}{*}{Method} & \multirow{2}{*}{Model}
& \multicolumn{2}{c}{Reference}
& \multicolumn{4}{c}{Noisy learning}
& \multicolumn{2}{c}{Noisy evaluation} \\
\cmidrule(lr){3-4}\cmidrule(lr){5-8}\cmidrule(l){9-10}
& & Initial & $C\!\to\!C$ & $N_{\mathrm{Inst}}\!\to\!C$ & $N_{\mathrm{Evid}}\!\to\!C$
& $N_{\mathrm{Rec}}\!\to\!C$ & $N_{\mathrm{Bind}}\!\to\!C$ & $C\!\to\!N_{\mathrm{Inst}}$ & $C\!\to\!N_{\mathrm{Evid}}$ \\
\midrule
\multicolumn{10}{@{}l}{\textbf{ALFWorld}} \\
\multirow{2}{*}{ACE} & DeepSeek-V4.1-Flash & \rsebase{95.42} & \rsebase{99.17} & \rsecell{97.92}{\downarrow}{1.3} & \rsecell{96.25}{\downarrow}{2.9} & \rsecell{100.00}{\uparrow}{0.8} & \rsecell{97.92}{\downarrow}{1.3} & \rseworst{93.75}{\downarrow}{5.5} & \rsecell{95.83}{\downarrow}{3.4} \\
 & GPT-5.2 & \rsebase{84.17} & \rsebase{96.25} & \rseworst{93.33}{\downarrow}{3.0} & \rsecell{96.67}{\uparrow}{0.4} & \rseworst{84.17}{\downarrow}{12.6} & \rsecell{92.08}{\downarrow}{4.3} & \rsecell{94.17}{\downarrow}{2.2} & \rsecell{90.00}{\downarrow}{6.5} \\
\multirow{2}{*}{RethinkSkill} & DeepSeek-V4.1-Flash & \rsebase{95.42} & \rsebase{99.17} & \rsecell{97.92}{\downarrow}{1.3} & \rsecell{97.08}{\downarrow}{2.1} & \rsecell{94.58}{\downarrow}{4.6} & \rsecell{100.00}{\uparrow}{0.8} & \rsecell{99.58}{\uparrow}{0.4} & \rsecell{100.00}{\uparrow}{0.8} \\
 & GPT-5.2 & \rsebase{84.17} & \rsebase{97.08} & \rsecell{94.17}{\downarrow}{3.0} & \rseworst{87.92}{\downarrow}{9.4} & \rsecell{85.83}{\downarrow}{11.6} & \rseworst{83.33}{\downarrow}{14.2} & \rsecell{93.75}{\downarrow}{3.4} & \rseworst{88.33}{\downarrow}{9.0} \\
\multirow{2}{*}{SkillOpt} & DeepSeek-V4.1-Flash & \rsebase{95.42} & \rsebase{98.33} & \rsecell{99.58}{\uparrow}{1.3} & \rsecell{96.25}{\downarrow}{2.1} & \rsecell{99.17}{\uparrow}{0.9} & \rsecell{99.58}{\uparrow}{1.3} & \rsecell{95.83}{\downarrow}{2.5} & \rsecell{97.92}{\downarrow}{0.4} \\
 & GPT-5.2 & \rsebase{84.17} & \rsebase{97.50} & \rsecell{94.58}{\downarrow}{3.0} & \rsecell{95.42}{\downarrow}{2.1} & \rsecell{95.00}{\downarrow}{2.6} & \rsecell{91.67}{\downarrow}{6.0} & \rsecell{92.92}{\downarrow}{4.7} & \rsecell{95.00}{\downarrow}{2.6} \\
\midrule
\multicolumn{10}{@{}l}{\textbf{SkillFlow}} \\
\multirow{2}{*}{ACE} & DeepSeek-V4.1-Flash & \rsebase{43.33} & \rsebase{44.67} & \rsecell{50.00}{\uparrow}{11.9} & \rsecell{42.67}{\downarrow}{4.5} & \rsecell{42.67}{\downarrow}{4.5} & \rsecell{41.33}{\downarrow}{7.5} & \rsecell{41.33}{\downarrow}{7.5} & \rsecell{34.67}{\downarrow}{22.4} \\
 & GPT-5.2 & \rsebase{32.67} & \rsebase{31.33} & \rsecell{34.00}{\uparrow}{8.5} & \rsecell{35.33}{\uparrow}{12.8} & \rsecell{29.33}{\downarrow}{6.4} & \rsecell{31.33}{\leftrightarrow}{0.0} & \rseworst{26.00}{\downarrow}{17.0} & \rseworst{17.33}{\downarrow}{44.7} \\
\multirow{2}{*}{RethinkSkill} & DeepSeek-V4.1-Flash & \rsebase{43.33} & \rsebase{46.00} & \rsecell{43.33}{\downarrow}{5.8} & \rsecell{38.67}{\downarrow}{15.9} & \rsecell{38.67}{\downarrow}{15.9} & \rsecell{38.00}{\downarrow}{17.4} & \rsecell{39.33}{\downarrow}{14.5} & \rsecell{44.00}{\downarrow}{4.3} \\
 & GPT-5.2 & \rsebase{32.67} & \rsebase{35.33} & \rsecell{31.33}{\downarrow}{11.3} & \rseworst{25.33}{\downarrow}{28.3} & \rsecell{38.00}{\uparrow}{7.6} & \rsecell{30.00}{\downarrow}{15.1} & \rsecell{32.67}{\downarrow}{7.5} & \rsecell{28.00}{\downarrow}{20.7} \\
\multirow{2}{*}{SkillOpt} & DeepSeek-V4.1-Flash & \rsebase{43.33} & \rsebase{63.33} & \rseworst{38.00}{\downarrow}{40.0} & \rsecell{56.00}{\downarrow}{11.6} & \rseworst{42.67}{\downarrow}{32.6} & \rsecell{44.67}{\downarrow}{29.5} & \rsecell{59.33}{\downarrow}{6.3} & \rsecell{61.33}{\downarrow}{3.2} \\
 & GPT-5.2 & \rsebase{32.67} & \rsebase{39.33} & \rsecell{34.00}{\downarrow}{13.6} & \rsecell{38.67}{\downarrow}{1.7} & \rsecell{35.33}{\downarrow}{10.2} & \rseworst{27.33}{\downarrow}{30.5} & \rsecell{36.00}{\downarrow}{8.5} & \rsecell{26.00}{\downarrow}{33.9} \\
\midrule
\multicolumn{10}{@{}l}{\textbf{ContinualSkillBench}} \\
\multirow{2}{*}{ACE} & DeepSeek-V4.1-Flash & \rsebase{49.17} & \rsebase{49.58} & \rsecell{52.08}{\uparrow}{5.0} & \rsecell{49.17}{\downarrow}{0.8} & \rsecell{50.42}{\uparrow}{1.7} & \rsecell{48.33}{\downarrow}{2.5} & \rsecell{49.58}{\leftrightarrow}{0.0} & \rsecell{48.75}{\downarrow}{1.7} \\
 & GPT-5.2 & \rsebase{50.00} & \rsebase{54.58} & \rseworst{46.67}{\downarrow}{14.5} & \rsecell{50.83}{\downarrow}{6.9} & \rseworst{48.75}{\downarrow}{10.7} & \rseworst{42.50}{\downarrow}{22.1} & \rsecell{53.33}{\downarrow}{2.3} & \rsecell{53.33}{\downarrow}{2.3} \\
\multirow{2}{*}{RethinkSkill} & DeepSeek-V4.1-Flash & \rsebase{49.17} & \rsebase{54.58} & \rsecell{50.83}{\downarrow}{6.9} & \rsecell{52.08}{\downarrow}{4.6} & \rsecell{49.17}{\downarrow}{9.9} & \rsecell{61.25}{\uparrow}{12.2} & \rseworst{47.92}{\downarrow}{12.2} & \rsecell{51.25}{\downarrow}{6.1} \\
 & GPT-5.2 & \rsebase{50.00} & \rsebase{56.67} & \rsecell{50.00}{\downarrow}{11.8} & \rsecell{52.50}{\downarrow}{7.4} & \rsecell{51.67}{\downarrow}{8.8} & \rsecell{54.58}{\downarrow}{3.7} & \rsecell{53.33}{\downarrow}{5.9} & \rsecell{53.33}{\downarrow}{5.9} \\
\multirow{2}{*}{SkillOpt} & DeepSeek-V4.1-Flash & \rsebase{49.17} & \rsebase{52.08} & \rsecell{60.00}{\uparrow}{15.2} & \rsecell{51.25}{\downarrow}{1.6} & \rsecell{64.17}{\uparrow}{23.2} & \rsecell{49.17}{\downarrow}{5.6} & \rsecell{60.00}{\uparrow}{15.2} & \rseworst{47.50}{\downarrow}{8.8} \\
 & GPT-5.2 & \rsebase{50.00} & \rsebase{58.75} & \rsecell{58.75}{\leftrightarrow}{0.0} & \rseworst{52.08}{\downarrow}{11.4} & \rsecell{53.75}{\downarrow}{8.5} & \rsecell{52.92}{\downarrow}{9.9} & \rsecell{60.83}{\uparrow}{3.5} & \rsecell{57.92}{\downarrow}{1.4} \\
\midrule
\multicolumn{10}{@{}l}{\textbf{RSE-Bench (Avg)}} \\
\multirow{2}{*}{ACE} & DeepSeek-V4.1-Flash & \rsebase{71.25} & \rsebase{74.53} & \rsecell{72.64}{\downarrow}{2.5} & \rsecell{69.27}{\downarrow}{7.1} & \rsecell{73.08}{\downarrow}{1.9} & \rsecell{71.19}{\downarrow}{4.5} & \rseworst{69.88}{\downarrow}{6.2} & \rsecell{69.67}{\downarrow}{6.5} \\
 & GPT-5.2 & \rsebase{61.42} & \rsebase{63.97} & \rsecell{63.17}{\downarrow}{1.3} & \rsecell{64.01}{\uparrow}{0.1} & \rseworst{59.68}{\downarrow}{6.7} & \rsecell{61.54}{\downarrow}{3.8} & \rsecell{61.76}{\downarrow}{3.4} & \rseworst{58.65}{\downarrow}{8.3} \\
\multirow{2}{*}{RethinkSkill} & DeepSeek-V4.1-Flash & \rsebase{71.25} & \rsebase{75.93} & \rsecell{72.92}{\downarrow}{4.0} & \rsecell{71.31}{\downarrow}{6.1} & \rsecell{71.10}{\downarrow}{6.4} & \rsecell{74.32}{\downarrow}{2.1} & \rsecell{72.04}{\downarrow}{5.1} & \rsecell{71.92}{\downarrow}{5.3} \\
 & GPT-5.2 & \rsebase{61.42} & \rsebase{66.93} & \rseworst{62.93}{\downarrow}{6.0} & \rseworst{60.40}{\downarrow}{9.7} & \rsecell{64.74}{\downarrow}{3.3} & \rseworst{61.53}{\downarrow}{8.1} & \rsecell{63.57}{\downarrow}{5.0} & \rsecell{61.96}{\downarrow}{7.4} \\
\multirow{2}{*}{SkillOpt} & DeepSeek-V4.1-Flash & \rsebase{71.25} & \rsebase{77.71} & \rsecell{74.60}{\downarrow}{4.0} & \rsecell{75.24}{\downarrow}{3.2} & \rsecell{75.79}{\downarrow}{2.5} & \rsecell{76.20}{\downarrow}{1.9} & \rsecell{77.74}{\uparrow}{<0.1} & \rsecell{74.81}{\downarrow}{3.7} \\
 & GPT-5.2 & \rsebase{61.42} & \rsebase{67.53} & \rsecell{63.79}{\downarrow}{5.5} & \rsecell{65.06}{\downarrow}{3.7} & \rsecell{63.46}{\downarrow}{6.0} & \rsecell{63.38}{\downarrow}{6.2} & \rsecell{66.49}{\downarrow}{1.5} & \rsecell{63.29}{\downarrow}{6.3} \\
\bottomrule
\end{tabular}%
}
\endgroup
\end{table*}

\begin{table}[t]
\centering
\caption{Noise conditions and data of RSE-Bench. N-Inst and N-Evid materialize a paired noisy variant of every task. N-Rec and N-Bind transform the learner's inputs at run time and add no items. $^\dagger$The clean-learning, clean-evaluation cell is a single baseline shared by all rows. Dashes mark undefined settings: held-out evaluation produces no learner experience.}
\label{tab:noise_cross}
\vspace{10pt}
\small
\setlength{\tabcolsep}{6pt}
\begin{tabular}{@{}llcccc@{}}
\toprule
 & & \multicolumn{2}{c}{Clean learning} & \multicolumn{2}{c}{Noisy learning} \\
\cmidrule(lr){3-4} \cmidrule(lr){5-6}
Entry point & Materialization & Clean eval & Noisy eval & Clean eval & Noisy eval \\
\midrule
N-Inst & data, all partitions & \checkmark$^\dagger$ & \checkmark & \checkmark & \checkmark \\
N-Evid & data, all partitions & \checkmark$^\dagger$ & \checkmark & \checkmark & \checkmark \\
N-Rec  & run-time seam        & \checkmark$^\dagger$ & --         & \checkmark & --         \\
N-Bind & run-time seam        & \checkmark$^\dagger$ & --         & \checkmark & --         \\
\bottomrule
\end{tabular}
\end{table}

\begin{table}[t]
\centering
\caption{Composition of RSE-Bench. Train, Sel., and Test denote the evolution-train, selection, and final-test partitions, respectively. Noise conditions follow the protocol in Table~\ref{tab:noise_cross}, with N-Rec and N-Bind applied only during learning. Sequential benchmarks (marked $^\dagger$) are split by stream position rather than random sampling; ALFWorld draws its partitions from the official train, valid-seen, and valid-unseen splits, respectively.}
\label{tab:data_statistics}
\small
\setlength{\tabcolsep}{5pt}
\begin{tabular}{llrrrr}
\toprule
Domain & Benchmark & Source pool & Train & Sel. & Test \\
\midrule
\multirow{1}{*}{Spreadsheet}
 & SpreadsheetBench & 400 & 50 & 30 & 80 \\
\midrule
\multirow{2}{*}{Document}
 & SP-DocVQA & 50{,}000 & 50 & 30 & 80 \\
 & MP-DocVQA & 46{,}436 & 50 & 30 & 80 \\
\midrule
\multirow{1}{*}{Interactive}
 & ALFWorld & 3{,}827 & 50 & 30 & 80 \\
\midrule
\multirow{2}{*}{Sequential$^\dagger$}
 & SkillFlow & 166 & 50 & 30 & 50 \\
 & ContinualSkillBench & 500 & 50 & 30 & 80 \\
\bottomrule
\end{tabular}
\end{table}

\begin{table*}[t]
\centering
\caption{Run-to-run variability on RSE-Bench. Scores (\%) are shown as $\mathrm{mean}_{\pm\mathrm{SD}}$ over three independent runs of the complete evolution and evaluation pipeline, with 50 held-out tasks for SkillFlow and 80 for each other benchmark. Columns specify learning $\rightarrow$ evaluation; $C$ is clean, and Inst, Evid, Rec, and Bind denote instruction, evidence, record, and binding noise. Underlines match the main table: within each benchmark and noise-condition column, they mark the largest relative decrease from the corresponding $C\!\to\!C$ baseline across the six method--model combinations, using unrounded rates and including exact ties.}
\label{tab:run_variability}
\vspace{10pt}
\begingroup
\fontsize{9}{10.4}\selectfont
\setlength{\tabcolsep}{1.6pt}
\renewcommand{\arraystretch}{1.12}
\newcommand{\rsecell}[2]{$#1_{\scalebox{0.78}{$\pm #2$}}$}
\newcommand{\rseworst}[2]{$\underline{#1}_{\scalebox{0.78}{$\pm #2$}}$}
\resizebox{\textwidth}{!}{%
\begin{tabular}{@{}ll*{8}{c}@{}}
\toprule
\multirow{2}{*}{Method} & \multirow{2}{*}{Model}
& \multicolumn{2}{c}{Reference}
& \multicolumn{4}{c}{Noisy learning}
& \multicolumn{2}{c}{Noisy evaluation} \\
\cmidrule(lr){3-4}\cmidrule(lr){5-8}\cmidrule(l){9-10}
& & Initial & $C\!\to\!C$ & $N_{\mathrm{Inst}}\!\to\!C$ & $N_{\mathrm{Evid}}\!\to\!C$
& $N_{\mathrm{Rec}}\!\to\!C$ & $N_{\mathrm{Bind}}\!\to\!C$ & $C\!\to\!N_{\mathrm{Inst}}$ & $C\!\to\!N_{\mathrm{Evid}}$ \\
\midrule
\multicolumn{10}{@{}l}{\textbf{SpreadsheetBench}} \\
\multirow{2}{*}{ACE} & DeepSeek-V4.1-Flash & \rsecell{79.58}{0.72} & \rsecell{90.00}{0.00} & \rsecell{74.58}{1.44} & \rseworst{74.17}{0.72} & \rsecell{83.33}{1.44} & \rsecell{80.00}{0.00} & \rseworst{77.50}{0.00} & \rseworst{78.75}{0.00} \\
 & GPT-5.2 & \rsecell{45.00}{2.17} & \rsecell{40.83}{0.72} & \rsecell{44.58}{0.72} & \rsecell{40.00}{2.17} & \rsecell{42.50}{3.31} & \rsecell{46.67}{1.44} & \rsecell{42.08}{1.44} & \rsecell{37.08}{1.44} \\
\multirow{2}{*}{RethinkSkill} & DeepSeek-V4.1-Flash & \rsecell{79.58}{0.72} & \rsecell{88.33}{0.72} & \rsecell{80.00}{1.25} & \rsecell{78.75}{0.00} & \rsecell{82.92}{0.72} & \rsecell{89.17}{3.15} & \rsecell{81.67}{0.72} & \rsecell{81.67}{1.91} \\
 & GPT-5.2 & \rsecell{45.00}{2.17} & \rsecell{45.83}{0.72} & \rsecell{39.58}{1.44} & \rsecell{40.00}{0.00} & \rsecell{52.08}{2.89} & \rseworst{39.17}{4.02} & \rsecell{44.58}{0.72} & \rsecell{45.00}{1.25} \\
\multirow{2}{*}{SkillOpt} & DeepSeek-V4.1-Flash & \rsecell{79.58}{0.72} & \rsecell{87.08}{3.15} & \rsecell{86.25}{0.00} & \rsecell{81.25}{1.25} & \rsecell{83.75}{1.25} & \rsecell{97.08}{1.44} & \rsecell{86.25}{1.25} & \rsecell{87.50}{1.25} \\
 & GPT-5.2 & \rsecell{45.00}{2.17} & \rsecell{47.92}{3.82} & \rseworst{33.75}{1.25} & \rsecell{46.25}{1.25} & \rseworst{39.58}{0.72} & \rsecell{45.42}{1.91} & \rsecell{50.42}{1.44} & \rsecell{47.08}{3.15} \\
\midrule
\multicolumn{10}{@{}l}{\textbf{SP-DocVQA}} \\
\multirow{2}{*}{ACE} & DeepSeek-V4.1-Flash & \rsecell{92.92}{1.44} & \rsecell{92.50}{0.00} & \rsecell{92.50}{0.00} & \rsecell{89.58}{1.91} & \rsecell{93.33}{2.89} & \rsecell{93.75}{2.17} & \rsecell{90.42}{1.91} & \rsecell{92.92}{1.44} \\
 & GPT-5.2 & \rsecell{88.33}{0.72} & \rsecell{90.83}{1.44} & \rsecell{91.25}{1.25} & \rsecell{89.58}{1.91} & \rseworst{85.00}{1.25} & \rsecell{89.17}{0.72} & \rsecell{86.25}{1.25} & \rsecell{88.33}{0.72} \\
\multirow{2}{*}{RethinkSkill} & DeepSeek-V4.1-Flash & \rsecell{92.92}{1.44} & \rsecell{97.92}{0.72} & \rsecell{97.50}{1.25} & \rsecell{96.25}{2.50} & \rsecell{96.25}{1.25} & \rseworst{93.75}{1.25} & \rsecell{94.17}{2.60} & \rseworst{91.67}{0.72} \\
 & GPT-5.2 & \rsecell{88.33}{0.72} & \rsecell{90.83}{1.44} & \rsecell{92.08}{1.91} & \rsecell{88.75}{0.00} & \rsecell{89.58}{1.91} & \rsecell{91.67}{0.72} & \rsecell{87.92}{0.72} & \rsecell{92.92}{1.44} \\
\multirow{2}{*}{SkillOpt} & DeepSeek-V4.1-Flash & \rsecell{92.92}{1.44} & \rsecell{96.25}{1.25} & \rsecell{96.67}{0.72} & \rsecell{96.67}{0.72} & \rsecell{96.67}{1.91} & \rsecell{95.42}{0.72} & \rsecell{95.42}{1.44} & \rsecell{95.42}{1.44} \\
 & GPT-5.2 & \rsecell{88.33}{0.72} & \rsecell{94.17}{1.44} & \rseworst{92.08}{3.15} & \rseworst{89.58}{1.91} & \rsecell{88.33}{1.91} & \rsecell{93.33}{1.44} & \rseworst{89.17}{1.91} & \rsecell{90.42}{0.72} \\
\midrule
\multicolumn{10}{@{}l}{\textbf{MP-DocVQA}} \\
\multirow{2}{*}{ACE} & DeepSeek-V4.1-Flash & \rsecell{67.08}{0.72} & \rsecell{71.25}{2.17} & \rsecell{68.75}{2.17} & \rseworst{63.75}{1.25} & \rsecell{68.75}{2.50} & \rsecell{65.83}{0.72} & \rsecell{66.67}{3.15} & \rsecell{67.08}{0.72} \\
 & GPT-5.2 & \rsecell{68.33}{2.60} & \rsecell{70.00}{1.25} & \rsecell{69.17}{1.44} & \rsecell{71.67}{1.91} & \rsecell{68.33}{0.72} & \rsecell{67.50}{1.25} & \rsecell{68.75}{0.00} & \rsecell{65.83}{1.91} \\
\multirow{2}{*}{RethinkSkill} & DeepSeek-V4.1-Flash & \rsecell{67.08}{0.72} & \rsecell{69.58}{1.44} & \rsecell{67.92}{0.72} & \rsecell{65.00}{1.25} & \rseworst{65.00}{0.00} & \rseworst{63.75}{1.25} & \rsecell{69.58}{1.91} & \rsecell{62.92}{0.72} \\
 & GPT-5.2 & \rsecell{68.33}{2.60} & \rsecell{75.83}{1.44} & \rseworst{70.42}{1.44} & \rsecell{67.92}{1.91} & \rsecell{71.25}{0.00} & \rsecell{70.42}{0.72} & \rseworst{69.17}{1.44} & \rseworst{64.17}{0.72} \\
\multirow{2}{*}{SkillOpt} & DeepSeek-V4.1-Flash & \rsecell{67.08}{0.72} & \rsecell{69.17}{0.72} & \rsecell{67.08}{2.60} & \rsecell{70.00}{0.00} & \rsecell{68.33}{0.72} & \rsecell{71.25}{1.25} & \rsecell{69.58}{1.91} & \rsecell{59.17}{0.72} \\
 & GPT-5.2 & \rsecell{68.33}{2.60} & \rsecell{67.50}{2.17} & \rsecell{69.58}{2.60} & \rsecell{68.33}{0.72} & \rsecell{68.75}{3.31} & \rsecell{69.58}{0.72} & \rsecell{69.58}{0.72} & \rsecell{63.33}{1.91} \\
\midrule
\multicolumn{10}{@{}l}{\textbf{ALFWorld}} \\
\multirow{2}{*}{ACE} & DeepSeek-V4.1-Flash & \rsecell{95.42}{0.72} & \rsecell{99.17}{0.72} & \rsecell{97.92}{0.72} & \rsecell{96.25}{2.50} & \rsecell{100.00}{0.00} & \rsecell{97.92}{0.72} & \rseworst{93.75}{1.25} & \rsecell{95.83}{1.44} \\
 & GPT-5.2 & \rsecell{84.17}{1.44} & \rsecell{96.25}{1.25} & \rseworst{93.33}{1.91} & \rsecell{96.67}{0.72} & \rseworst{84.17}{3.61} & \rsecell{92.08}{0.72} & \rsecell{94.17}{3.61} & \rsecell{90.00}{2.50} \\
\multirow{2}{*}{RethinkSkill} & DeepSeek-V4.1-Flash & \rsecell{95.42}{0.72} & \rsecell{99.17}{0.72} & \rsecell{97.92}{0.72} & \rsecell{97.08}{0.72} & \rsecell{94.58}{0.72} & \rsecell{100.00}{0.00} & \rsecell{99.58}{0.72} & \rsecell{100.00}{0.00} \\
 & GPT-5.2 & \rsecell{84.17}{1.44} & \rsecell{97.08}{2.60} & \rsecell{94.17}{0.72} & \rseworst{87.92}{0.72} & \rsecell{85.83}{0.72} & \rseworst{83.33}{0.72} & \rsecell{93.75}{2.50} & \rseworst{88.33}{1.44} \\
\multirow{2}{*}{SkillOpt} & DeepSeek-V4.1-Flash & \rsecell{95.42}{0.72} & \rsecell{98.33}{1.44} & \rsecell{99.58}{0.72} & \rsecell{96.25}{1.25} & \rsecell{99.17}{0.72} & \rsecell{99.58}{0.72} & \rsecell{95.83}{1.91} & \rsecell{97.92}{1.44} \\
 & GPT-5.2 & \rsecell{84.17}{1.44} & \rsecell{97.50}{0.00} & \rsecell{94.58}{0.72} & \rsecell{95.42}{0.72} & \rsecell{95.00}{2.50} & \rsecell{91.67}{0.72} & \rsecell{92.92}{0.72} & \rsecell{95.00}{0.00} \\
\midrule
\multicolumn{10}{@{}l}{\textbf{SkillFlow}} \\
\multirow{2}{*}{ACE} & DeepSeek-V4.1-Flash & \rsecell{43.33}{1.15} & \rsecell{44.67}{3.06} & \rsecell{50.00}{0.00} & \rsecell{42.67}{1.15} & \rsecell{42.67}{2.31} & \rsecell{41.33}{1.15} & \rsecell{41.33}{1.15} & \rsecell{34.67}{1.15} \\
 & GPT-5.2 & \rsecell{32.67}{2.31} & \rsecell{31.33}{1.15} & \rsecell{34.00}{0.00} & \rsecell{35.33}{2.31} & \rsecell{29.33}{3.06} & \rsecell{31.33}{1.15} & \rseworst{26.00}{2.00} & \rseworst{17.33}{1.15} \\
\multirow{2}{*}{RethinkSkill} & DeepSeek-V4.1-Flash & \rsecell{43.33}{1.15} & \rsecell{46.00}{2.00} & \rsecell{43.33}{1.15} & \rsecell{38.67}{1.15} & \rsecell{38.67}{1.15} & \rsecell{38.00}{2.00} & \rsecell{39.33}{1.15} & \rsecell{44.00}{2.00} \\
 & GPT-5.2 & \rsecell{32.67}{2.31} & \rsecell{35.33}{1.15} & \rsecell{31.33}{1.15} & \rseworst{25.33}{1.15} & \rsecell{38.00}{2.00} & \rsecell{30.00}{0.00} & \rsecell{32.67}{3.06} & \rsecell{28.00}{0.00} \\
\multirow{2}{*}{SkillOpt} & DeepSeek-V4.1-Flash & \rsecell{43.33}{1.15} & \rsecell{63.33}{1.15} & \rseworst{38.00}{0.00} & \rsecell{56.00}{2.00} & \rseworst{42.67}{1.15} & \rsecell{44.67}{1.15} & \rsecell{59.33}{1.15} & \rsecell{61.33}{1.15} \\
 & GPT-5.2 & \rsecell{32.67}{2.31} & \rsecell{39.33}{2.31} & \rsecell{34.00}{2.00} & \rsecell{38.67}{1.15} & \rsecell{35.33}{2.31} & \rseworst{27.33}{2.31} & \rsecell{36.00}{0.00} & \rsecell{26.00}{0.00} \\
\midrule
\multicolumn{10}{@{}l}{\textbf{ContinualSkillBench}} \\
\multirow{2}{*}{ACE} & DeepSeek-V4.1-Flash & \rsecell{49.17}{1.44} & \rsecell{49.58}{0.72} & \rsecell{52.08}{1.91} & \rsecell{49.17}{0.72} & \rsecell{50.42}{0.72} & \rsecell{48.33}{0.72} & \rsecell{49.58}{4.02} & \rsecell{48.75}{5.45} \\
 & GPT-5.2 & \rsecell{50.00}{0.00} & \rsecell{54.58}{1.91} & \rseworst{46.67}{0.72} & \rsecell{50.83}{1.44} & \rseworst{48.75}{1.25} & \rseworst{42.50}{1.25} & \rsecell{53.33}{0.72} & \rsecell{53.33}{0.72} \\
\multirow{2}{*}{RethinkSkill} & DeepSeek-V4.1-Flash & \rsecell{49.17}{1.44} & \rsecell{54.58}{0.72} & \rsecell{50.83}{1.91} & \rsecell{52.08}{0.72} & \rsecell{49.17}{1.91} & \rsecell{61.25}{1.25} & \rseworst{47.92}{0.72} & \rsecell{51.25}{0.00} \\
 & GPT-5.2 & \rsecell{50.00}{0.00} & \rsecell{56.67}{1.44} & \rsecell{50.00}{1.25} & \rsecell{52.50}{2.17} & \rsecell{51.67}{0.72} & \rsecell{54.58}{1.44} & \rsecell{53.33}{1.91} & \rsecell{53.33}{0.72} \\
\multirow{2}{*}{SkillOpt} & DeepSeek-V4.1-Flash & \rsecell{49.17}{1.44} & \rsecell{52.08}{1.44} & \rsecell{60.00}{1.25} & \rsecell{51.25}{3.31} & \rsecell{64.17}{0.72} & \rsecell{49.17}{1.91} & \rsecell{60.00}{1.25} & \rseworst{47.50}{1.25} \\
 & GPT-5.2 & \rsecell{50.00}{0.00} & \rsecell{58.75}{0.00} & \rsecell{58.75}{0.00} & \rseworst{52.08}{3.15} & \rsecell{53.75}{1.25} & \rsecell{52.92}{1.91} & \rsecell{60.83}{1.91} & \rsecell{57.92}{0.72} \\
\bottomrule
\end{tabular}%
}
\endgroup
\end{table*}

\begin{table*}[t]
\centering
\caption{Crossing learning and evaluation conditions for ACE with DeepSeek-V4.1-Flash on SpreadsheetBench. Scores are percentages, and Initial denotes the agent without an evolved skill store. Under $N_{\mathrm{Inst}}$ and $N_{\mathrm{Evid}}$ evaluation, the cleanly evolved agent scores $6.25$ and $3.75$ percentage points below the initial agent, respectively. Learning under the matching noise condition yields scores $6.25$ and $8.75$ percentage points below clean learning under the same noisy evaluation conditions.}
\label{tab:cross_noise_spreadsheet}
\vspace{10pt}
\begingroup
\fontsize{9}{10.4}\selectfont
\setlength{\tabcolsep}{1.6pt}
\renewcommand{\arraystretch}{1.12}
\newcommand{\rsecell}[2]{$#1_{\scalebox{0.78}{$\pm #2$}}$}
\newcommand{\rseworst}[2]{$\underline{#1}_{\scalebox{0.78}{$\pm #2$}}$}
\resizebox{\textwidth}{!}{%
\begin{tabular}{@{}*{12}{c}@{}}
\toprule
\multicolumn{4}{c}{Reference}
& \multicolumn{4}{c}{Noisy learning}
& \multicolumn{4}{c}{Noisy evaluation} \\
\cmidrule(lr){1-4}\cmidrule(lr){5-8}\cmidrule(l){9-12}
$\mathrm{Initial}\!\to\!C$ & $C\!\to\!C$
& $\mathrm{Initial}\!\to\!N_{\mathrm{Inst}}$
& $\mathrm{Initial}\!\to\!N_{\mathrm{Evid}}$
& $N_{\mathrm{Inst}}\!\to\!C$
& $N_{\mathrm{Evid}}\!\to\!C$
& $N_{\mathrm{Rec}}\!\to\!C$
& $N_{\mathrm{Bind}}\!\to\!C$
& $C\!\to\!N_{\mathrm{Inst}}$
& $C\!\to\!N_{\mathrm{Evid}}$
& $N_{\mathrm{Inst}}\!\to\!N_{\mathrm{Inst}}$
& $N_{\mathrm{Evid}}\!\to\!N_{\mathrm{Evid}}$ \\
\midrule
79.58 & 90.00
& 83.75 & 82.50
& 74.58 & 74.17
& 83.33 & 80.00
& 77.50 & 78.75
& 71.25 & 70.00 \\
\bottomrule
\end{tabular}%
}
\endgroup
\end{table*}

\begin{table*}[t]
\centering
\caption{Ablation on training rounds
(ACE, DeepSeek-V4.1-Flash, SpreadsheetBench; scores in \%).
Results are from one run per learning condition.
Each column reports performance on clean held-out tasks using the
skill checkpoint at the end of round $R$.}
\label{tab:training_stages}
\vspace{10pt}
\begin{tabular}{lccc}
\toprule
Training data & $R=1$ & $R=2$ & $R=3$ \\
\midrule
Clean & 85.00 & 86.25 & 90.00 \\
$N_2$ & 76.25 & 75.00 & 73.75 \\
$\Delta$ ($N_2-\mathrm{Clean}$, pp) & $-8.75$ & $-11.25$ & $-16.25$ \\
\bottomrule
\end{tabular}
\end{table*}

\begin{table*}[t]
\centering
\caption{Backbone ablation with GPT-5.5 for ACE on SpreadsheetBench. Scores are percentages. Subscripts report relative changes, $100(s-b)/b$, from the clean-learning, clean-evaluation baseline $b=95.00$. Among the four learning-side noise conditions, $N_{\mathrm{Evid}}$ produces the largest score decrease.}
\label{tab:backbone_gpt55}
\begingroup
\fontsize{9}{10.4}\selectfont
\setlength{\tabcolsep}{1.6pt}
\renewcommand{\arraystretch}{1.12}
\resizebox{\textwidth}{!}{%
\begin{tabular}{@{}*{8}{c}@{}}
\toprule
\multicolumn{2}{c}{Reference}
& \multicolumn{4}{c}{Noisy learning}
& \multicolumn{2}{c}{Noisy evaluation} \\
\cmidrule(lr){1-2}\cmidrule(lr){3-6}\cmidrule(l){7-8}
Initial & $C\!\to\!C$
& $N_{\mathrm{Inst}}\!\to\!C$
& $N_{\mathrm{Evid}}\!\to\!C$
& $N_{\mathrm{Rec}}\!\to\!C$
& $N_{\mathrm{Bind}}\!\to\!C$
& $C\!\to\!N_{\mathrm{Inst}}$
& $C\!\to\!N_{\mathrm{Evid}}$ \\
\midrule
$88.75$ & $95.00$
& $93.75_{\downarrow 1.3\%}$
& $88.75_{\downarrow 6.6\%}$
& $90.00_{\downarrow 5.3\%}$
& $91.25_{\downarrow 3.9\%}$
& $86.25_{\downarrow 9.2\%}$
& $88.75_{\downarrow 6.6\%}$ \\
\bottomrule
\end{tabular}%
}
\endgroup
\end{table*}


\end{document}